\documentclass[11pt]{article}

\usepackage[final]{acl}

\usepackage{times}
\usepackage{latexsym}

\usepackage[T1]{fontenc}

\usepackage[utf8]{inputenc}

\usepackage{microtype}

\usepackage{inconsolata}

\usepackage{graphicx}

\usepackage{amsmath}

\usepackage{booktabs}
\usepackage{multirow}
\usepackage{array}
\usepackage{pifont}
\usepackage{xcolor}
\usepackage{amssymb}
\usepackage{tcolorbox}
\tcbuselibrary{breakable, skins}
\usepackage{fontawesome5}
\usepackage[export]{adjustbox}
\usepackage{bbding}
\usepackage{enumitem}
\usepackage{listings}

\definecolor{softpurple}{HTML}{E6CEE3}
\definecolor{softblue}{HTML}{E1E1F9}
\definecolor{softorange}{HTML}{FCD5B5}

\newcommand{\hlfor}[1]{\setlength{\fboxsep}{1.5pt}\colorbox{softpurple}{#1}}
\newcommand{\hlcons}[1]{\setlength{\fboxsep}{1.5pt}\colorbox{softblue}{#1}}
\newcommand{\hllogic}[1]{\setlength{\fboxsep}{1.5pt}\colorbox{softorange}{#1}}

\definecolor{deepgreen}{RGB}{6,153,6} 
\definecolor{deepred}{RGB}{254,34,35}  
\definecolor{darkyellow}{RGB}{188, 158, 0}

\newcommand{\halfcheck}{%
    \textcolor{darkyellow}{\CheckmarkBold\hspace{-0.305cm}\XSolidBrush}%
}
\newcommand{\halfchecktext}{
    \textcolor{darkyellow}{\CheckmarkBold\hspace{-0.345cm}\XSolidBrush}
}

\newcommand{\tabcrossmark}{%
    \raisebox{-0.5ex}{\textcolor{deepred}{\XSolidBrush}}%
}

\newcommand{\tabcheckmark}{%
    \raisebox{-0.5ex}{\textcolor{deepgreen}{\CheckmarkBold}}%
}

\newtcolorbox{promptbox}[1][]{
  colback=gray!5!white,      
  colframe=gray!75!black,    
  title=\textbf{System Prompt}, 
  coltitle=white,            
  fonttitle=\bfseries\sffamily,
  enhanced,
  boxrule=0.5mm,
  left=5pt, right=5pt, top=5pt, bottom=5pt,
  fontupper=\ttfamily\small\upshape, 
  #1
}

\lstdefinelanguage{json}{
    basicstyle=\footnotesize\ttfamily,
    numbers=none,
    frame=lines,      
    framesep=2mm,     
    breaklines=true,  
    captionpos=b,     
    keepspaces=true,
    columns=flexible,
    showstringspaces=false,
    commentstyle=\color{gray},
    keywordstyle=\color{blue},
    stringstyle=\color{black!70},
    keywords={true, false, null} 
}

\title{Towards Generalizable Visually Grounded Exploration of Household Devices}

\author{
 \textbf{Linhao Zheng\textsuperscript{1\footnotemark[1]}},
 \textbf{Zeming Liu\textsuperscript{2\footnotemark[1]}},
 \textbf{Wangke Chen\textsuperscript{1}},
 \textbf{Li Zeng\textsuperscript{1}},
 \\
 \textbf{Wanxiang Che\textsuperscript{3}},
 \textbf{Heyan Huang\textsuperscript{1}},
 \textbf{Yuhang Guo\textsuperscript{1\footnotemark[2]}}
\\
 \textsuperscript{1}School of Computer Science and Technology, Beijing Institute of Technology
\\
 \textsuperscript{2}School of Computer Science and Engineering, Beihang University
\\
 \textsuperscript{3}Research Center for Social Computing and Interactive Robotics, Harbin Institute of Technology
\\
 \href{zhenglinhao@bit.edu.cn}{\textcolor{black}{zhenglinhao}}@bit.edu.cn
 \href{zmliu@buaa.edu.cn}{\textcolor{black}{zmliu}}@buaa.edu.cn
 \href{guoyuhang@bit.edu.cn}{\textcolor{black}{guoyuhang}}@bit.edu.cn
}

\begin{document}
\maketitle

\begin{abstract}
\renewcommand{\thefootnote}{\fnsymbol{footnote}} 
\footnotetext[1]{Equal contribution}
\footnotetext[2]{Corresponding author: \href{guoyuhang@bit.edu.cn}{guoyuhang@bit.edu.cn}}
\renewcommand{\thefootnote}{\arabic{footnote}}
Recent advancements in Vision-Language Models (VLMs) have demonstrated impressive capabilities in static visual recognition and high-level semantic reasoning. However, current embodied exploration paradigms still heavily rely on imitation learning from human-annotated trajectories, which severely limits agents’ generalization ability. The key bottleneck of realizing general autonomous embodied agents lies in \textit{Generalizable Visually Grounded Exploration}—the ability to operate novel devices without manuals or specific training by actively grounding abstract world knowledge into fine-grained visual affordances. Yet, existing benchmarks fail to evaluate this capability: they generally rely on explicit documents and annotated trajectories, neglecting the dynamic \textit{Hypothesis-Interaction-Refinement} process essential for functional device operation.
To bridge this gap, we introduce \textbf{VGEBench}, a comprehensive benchmark designed to evaluate the generalizable visually grounded exploration capabilities of VLMs. Unlike static datasets, we construct a Logic-Driven State Machine framework. This framework simulates multi-turn interaction loops, compelling agents to achieve goals by active visual perception and feedback-driven correction. Experimental results demonstrate that existing VLMs face significant challenges in translating semantic knowledge into physical execution and maintaining long-horizon state tracking.
\footnote{Dataset and codes are publicly available at \url{https://github.com/BITHLP/VGEBench}}
\end{abstract}

\section{Introduction}

\begin{figure}[t!]
  \centering
  \includegraphics[width=\linewidth]{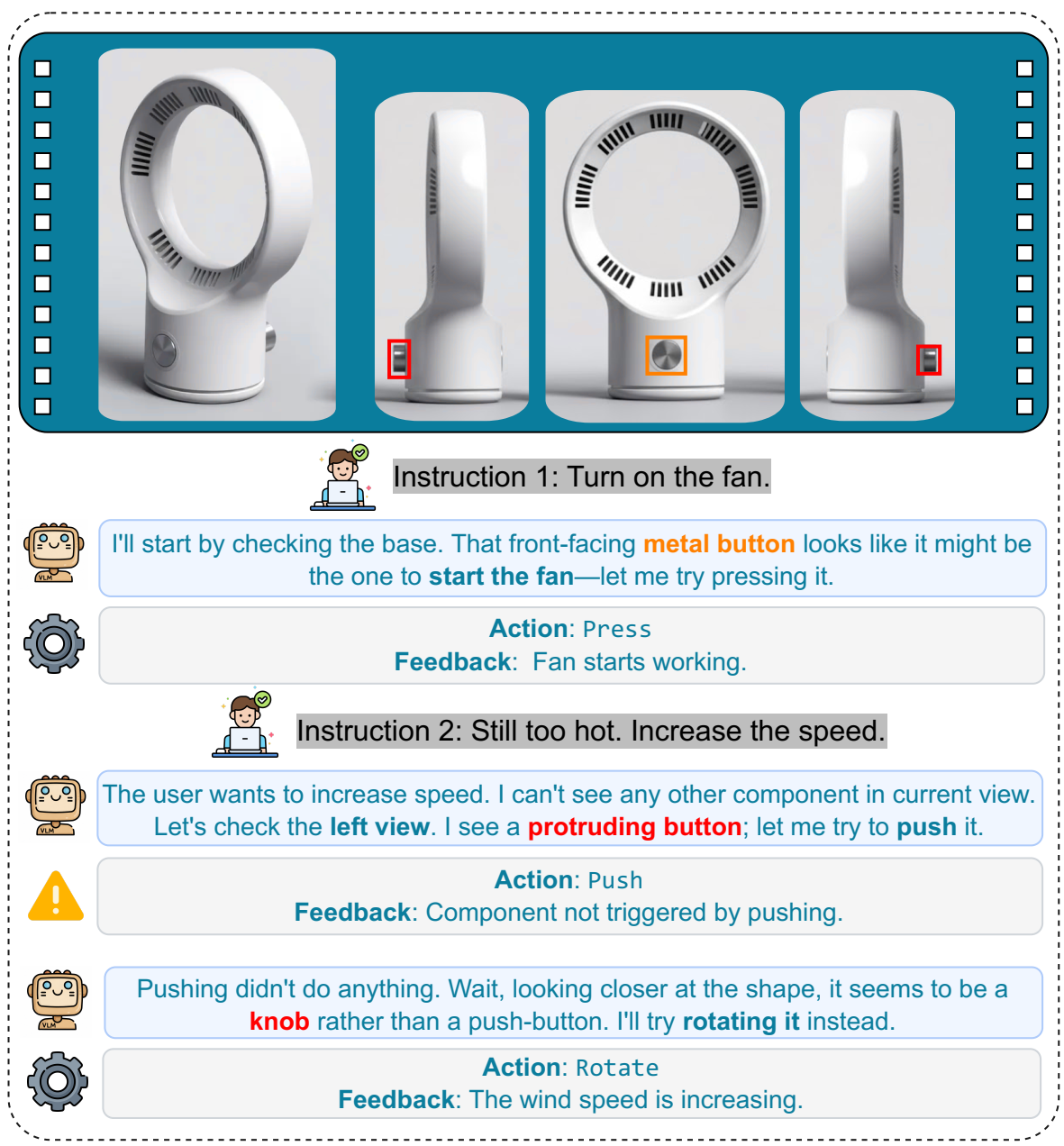}
  \caption{An example of \textbf{Generalizable Visually Grounded Exploration}. The example demonstrates a \textbf{Hypothesis-Interaction-Refinement} loop: after failing to increase the speed by pushing a component, the agent interprets the physical feedback, updates its hypothesis, re-evaluates the visual features to identify the component as a rotatable knob, and successfully corrects its action.}
  \label{fig:intro_fig}
\end{figure}

The ultimate goal of Artificial General Intelligence (AGI) is to develop embodied agents capable of assisting humans by seamlessly interacting with the devices in their daily lives~\cite{duan2022surveyembodiedaisimulators, wang2023voyager, driess2023palme}. From heating food in a microwave to adjusting a standing fan, these activities constitute the fundamental fabric of daily living. While humans can effortlessly manipulate a never-before-seen device without consulting a manual, this ability remains a formidable challenge for current AI systems. Humans achieve this proficiency not by memorizing the operation of every specific device instance, but through \textit{generalizable exploration} grounded in general world knowledge. When encountering a novel device, a human forms a hypothesis based on affordance perception (e.g., ``the \faSnowflake \hspace{0.6ex}symbol might represent temperature control'')~\cite{qian2024affordancellm}, interacts with it, and refines their understanding based on physical feedback (e.g., ``turning it left didn't work, so I must turn it right'')~\cite{bohg2017interactive}. This dynamic process of \textbf{Hypothesis-Interaction-Refinement} is the cornerstone of generalizable manipulation. 

However, current research in embodied AI and Large Language Models (LLMs) has yet to fully capture this capability. On one hand, traditional robotic learning benchmarks typically rely on Imitation Learning or Reinforcement Learning~\cite{tao2024maniskill3, li2024behavior1k, ALFWorld20}, requiring agents to be trained on specific environments, which limits their ability of generalization to unseen devices~\cite{liang2022code}. On the other hand, LLM-based tool learning~\cite{schick2023toolformer, qin2023toolllm} has achieved success in software domains, but primarily operates on structured APIs with explicit documentation. These paradigms bypass the core challenge of the physical world: the need to bridge the gap between abstract common sense and instance-specific physical interfaces through active visual exploration and feedback reasoning. 

To bridge this gap, we introduce \textbf{VGEBench}, a novel benchmark designed to evaluate the \textit{Generalizable Visually Grounded Exploration} capabilities of Vision Language Models (VLMs) on household devices. Unlike previous benchmarks that focus on static visual recognition or rote execution, VGEBench simulates a high-fidelity, documentation-free environment where agents must figure out how to use 968 diverse devices across 26 categories solely through visual perception and interaction. Our setting poses unique challenges: (1) \textbf{Knowledge Grounding}: The agent must map abstract knowledge (e.g., ``fans oscillate'') to fine-grained, unlabelled visual affordances (e.g., a specific mechanical pin); (2)\textbf{Visual Grounding}: The agent must transcend static recognition to precisely localize interactive components, serving as the spatial prerequisite for valid physical interaction; (3) \textbf{Feedback-Driven Reasoning}: The agent must interpret diverse environmental feedback—ranging from state changes (lights turning on) to failures (pressing a knob)—to correct its plans in real-time. 

Our main contributions are summarized as follows: 
\begin{itemize} 
  \item We identify \textit{Generalizable Visually Grounded Exploration} as a critical missing link for VLM agents in operating household devices, revealing a critical gap between abstract world knowledge and instance-level affordances.
  \item To bridge this gap, we present VGEBench, a novel benchmark which features a Logic-Driven State Machine framework that supports deterministic, multi-turn interactions, enabling the evaluation of long-horizon generalizable exploratory reasoning.
  \item We establish a multi-dimensional evaluation protocol and conduct extensive experiments on various VLMs, revealing their limitations in visual grounding, generalizable exploration, and long-horizon state tracking. Our results show that existing models struggle to translate general world knowledge into precise physical interaction in multi-turn interaction scenarios.
\end{itemize}

\section{Related Work}

\begin{table*}[t!]
\centering
\small
\setlength{\tabcolsep}{4pt}
\renewcommand{\arraystretch}{1.1}

\begin{tabular}{l ccc cc ccc c}
\toprule
\multirow{2}{*}{\textbf{Benchmark}} & \multicolumn{3}{c}{\textbf{Exploration Capabilities}} & \multicolumn{2}{c}{\textbf{Functional Logic}} & \multicolumn{3}{c}{\textbf{Vision}} & \multirow{2}{*}{\textbf{Domain}} \\
\cmidrule(lr){2-4} \cmidrule(lr){5-6} \cmidrule(lr){7-9}
 & \textbf{\makebox[0.7cm]{GE}} & \textbf{\makebox[0.7cm]{IFR}} & \textbf{\makebox[0.7cm]{EC}} & \textbf{DLM} & \textbf{MF} & \textbf{PE} & \textbf{FG} & \textbf{MV} & \\
\midrule

BFCL~\cite{patil2025the} 
& \tabcheckmark & \tabcheckmark & \tabcrossmark & \tabcrossmark & \tabcrossmark & \tabcrossmark & \tabcrossmark & \tabcrossmark & API/Code \\

PhysToolBench~\cite{zhang2025phystoolbench} 
& \tabcheckmark & \tabcrossmark & \tabcrossmark & \tabcrossmark & \tabcheckmark & \tabcheckmark & \tabcrossmark & \tabcrossmark & Physical Tools \\

OpenEQA~\cite{majumdar2024openeqa} 
& \tabcheckmark & \halfcheck & \tabcrossmark & \tabcrossmark & \tabcheckmark & \tabcheckmark & \tabcrossmark & \tabcheckmark & Indoor Scenes \\

OSWorld~\cite{xie2024osworld}
& \tabcheckmark & \halfcheck & \tabcheckmark & \tabcheckmark & \halfcheck & \tabcrossmark & \tabcheckmark & \tabcrossmark & Computer GUIs \\

AndroidWorld~\cite{rawles2024androidworld}
& \tabcheckmark & \tabcheckmark & \tabcheckmark & \tabcheckmark & \tabcheckmark & \tabcrossmark & \tabcheckmark & \tabcrossmark & Android Apps \\

HomeBench~\cite{li2025homebench} 
& \tabcrossmark & \tabcrossmark & \tabcrossmark & \tabcheckmark & \tabcrossmark & \tabcrossmark & \tabcrossmark & \tabcrossmark & Virtual Smart Home \\

RoboCasa~\cite{robocasa2024} 
& \tabcrossmark & \tabcrossmark & \tabcheckmark & \tabcrossmark & \tabcheckmark & \tabcheckmark & \tabcrossmark & \tabcheckmark & Kitchen Tasks \\

BEHAVIOR-1k~\cite{li2024behavior1k} 
& \tabcrossmark & \tabcheckmark & \tabcheckmark & \tabcrossmark & \tabcheckmark & \tabcheckmark & \tabcrossmark & \tabcheckmark & Household Tasks \\

\midrule
\textbf{VGEBench (Ours)}
& \tabcheckmark & \tabcheckmark & \tabcheckmark & \tabcheckmark & \tabcheckmark & \tabcheckmark & \tabcheckmark & \tabcheckmark & Household Devices \\

\bottomrule
\end{tabular}

\caption{
Comparison of VGEBench with existing related benchmarks.
\textbf{Exploration:}
\textbf{GE (Generalizable Exploration)}: training-free generalization without environment-specific RL/IL finetuning;
\textbf{IFR (Interactive Feedback \& Refinement)}: Supporting multi-turn hypothesis testing;
\textbf{EC (Error Correction)}: Environment can support multi-turn trial-and-error.
\textbf{Logic:}
\textbf{DLM (Device Logic Modeling)}: Providing explicit device state and transition logic;
\textbf{MF (Manual-Free)}: Relying on world knowledge and environmental feedback rather than provided documentation.
\textbf{Vision:}
\textbf{PE (Physical Entity)}: Targeting 3D functional appliances with realistic topology;
\textbf{FG (Fine-grained Grounding)}: Localizing specific interactive parts (e.g., buttons) vs. coarse objects;
\textbf{MV (Multi-View)}: Active perception across multiple viewpoints.
\tabcheckmark: Fully Supported; \halfchecktext: Partially Supported; \tabcrossmark: Not Supported.
}
\label{tab:related_benchmark}
\end{table*}

\paragraph{From API to Physical Interaction.} 
Large language models have demonstrated strong tool-use capability in language-centric settings, where tools are accessed through structured API interfaces~\cite{schick2023toolformer,qin2023toolllm}. While advanced reasoning frameworks enhance decision-making through iterative thought traces~\cite{yao2023react, shinn2023reflexion}, these agents largely operate under a \textit{manual-dependent} paradigm~\cite{patil2025the, li2025homebench}, relying on explicit documentation or schema definitions provided in the prompt rather than discovering functionality through exploration~\cite{cheng2025toolspectrum}.
Moving to operating systems, agents are required to navigate GUI environments~\cite{xie2024osworld, rawles2024androidworld}. While they support open-ended tasks, their perception is often simplified by accessing the accessibility tree (e.g., DOM/XML), and their interaction is confined to 2D screen spaces, lacking the \textit{multi-view} active perception required in the 3D physical world.

\paragraph{Fine-grained Visual Grounding.}
Operating physical devices not only requires an understanding of the device's functionality but hinges on fine-grained visual grounding of interactive components (e.g., knobs, ports, or mode buttons). Traditional VQA-style benchmarks have advanced holistic vision-language understanding~\cite{antol2015vqa,hudson2019gqa,marino2019ok} and physical object reasoning~\cite{ma2025phyblock, zhang2025phystoolbench}, yet they largely evaluate \textit{static} question answering, without explicitly modeling actionable component localization under interaction constraints~\cite{you2023ferret}.
Although there has been progress on linking language to specific regions or instances through referring expression comprehension~\cite{yu2016modeling, liu2024grounding} and grounded multimodal interfaces~\cite{peng2023kosmos, chen2023shikra}, there remains a practical granularity mismatch in current VLMs: models can be robust in holistic recognition but brittle when correct actions depend on small, functionally meaningful components that are easy to confuse across instances and viewpoints~\cite{geng2022gapartnet}.

\paragraph{Embodied Exploration.}
Beyond knowing where to act, operating household devices requires anticipating what will happen when acting, with actions ultimately validated through observable state changes and feedback. Recently, benchmarks like OpenEQA~\cite{majumdar2024openeqa} and PEAP~\cite{lan2026peap} have focused on assessing embodied perception. However, they primarily evaluate passive observation or static functional reasoning rather than active, state-changing manipulation.
In physical environments, traditional benchmarks (e.g., ALFWorld~\cite{ALFWorld20}, BEHAVIOR~\cite{li2024behavior1k}) have established high-fidelity simulation environments for common household tasks. However, they typically rely on training agents within \textbf{specific simulators}—often via \textbf{Imitation Learning} on extensive \textbf{human action trajectories}~\cite{savva2019habitat}. This paradigm inherently limits their generalizability to novel instances outside the training distribution. 
The central challenge remains \textbf{generalizable exploration}: enabling agents to map abstract world knowledge to instance-specific affordances without prior training or manuals. Unlike previous works that focus on either text-heavy reasoning or rote skill execution, VGEBench targets the intersection of visual grounding, active logic probing, and feedback-driven refinement.

\section{VGEBench}
\label{sec:dataset}

\begin{figure*}[t!]
  \centering
  \includegraphics[width=\linewidth]{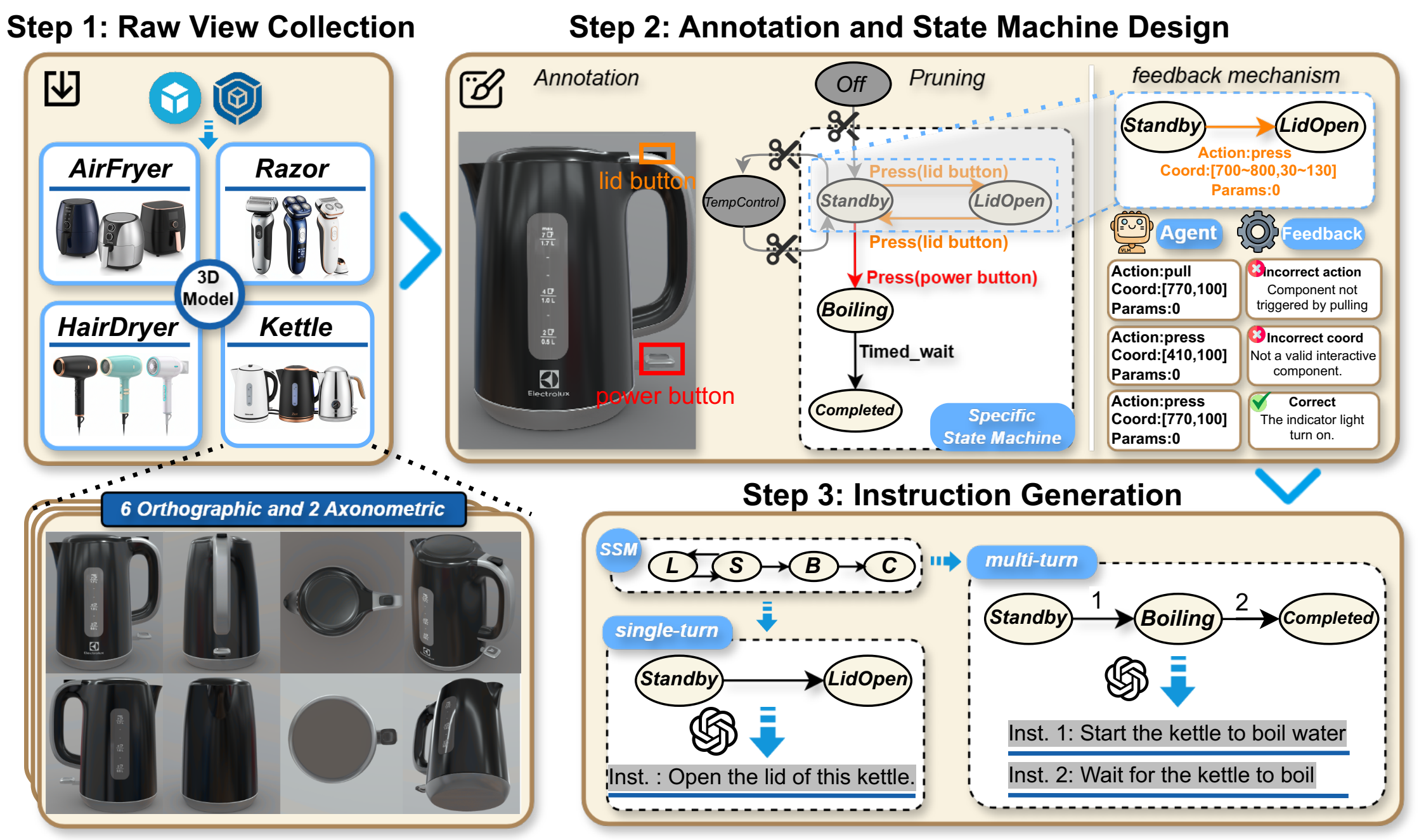}
  \caption{\textbf{The Data Construction Pipeline of VGEBench:} This pipeline consists of three main phases: raw view collection, Annotation and state machine design, and instruction generation. In the \textbf{Raw View Collection} phase, we acquire high-fidelity visual observations, capturing views for each 3D object to ensure comprehensive coverage. The process then proceeds to \textbf{Annotation and State Machine Design}, where a Universal Category State Machine (UCSM) is instantiated into a Specific State Machine (SSM) based on manual annotations of visible interactive components. Building on this SSM, our environment simulator can generate state-dependent feedbacks. Finally, \textbf{Instruction Generation} synthesizes diverse tasks by sampling valid state transition paths from the SSM and refining them into natural language instructions via LLMs.}
  \label{fig:data_pipeline_fig}
\end{figure*}

To accurately evaluate the generalizable visually grounded exploration capabilities of VLMs in household environments, we constructed VGEBench, a large-scale benchmark of diverse household devices with fine-grained component annotations and executable interaction logic. As shown in Fig.~\ref{fig:data_pipeline_fig}, the data construction pipeline consists of three main stages: raw view collection, state machine annotation, and instruction generation.

\subsection{Raw View Collection}
To accurately evaluate fine-grained manipulation capabilities (e.g., rotating a specific knob or toggling a small switch), high-fidelity visual observations are essential. Traditional 3D model datasets (e.g., ShapeNetCore~\cite{shapenet2015}) often lack the texture resolution or distinct functional details required for such precise interactions. Motivated by this gap, we sourced high-quality 3D models that support legible component identification from Sketchfab\footnote{\url{https://sketchfab.com/}} and 3D Warehouse\footnote{\url{https://3dwarehouse.sketchup.com/}}. Utilizing the platforms' built-in 360-degree model preview features, human annotators captured 8 multi-view images per object (6 orthographic and 2 axonometric). We also conducted a rigorous manual verification for every candidate model to ensure the presence of clearly identifiable interactive components. We finally collected 968 high-fidelity 3D models across 26 common categories (e.g., CoffeeMachine, DigitalAlarmClock, WashingMachine), totaling 7,888 collected views. A detailed distribution is provided in Tab.~\ref{tab:category_distribution}.

\subsection{Annotation and State Machine Design}
We propose a logic-driven annotation protocol to ground visual perception into executable plans. First, human annotators labeled the bounding boxes of all \textit{interactive components} (e.g., buttons, knobs) visible across the collected views. To govern the interaction logic, we designed a \textbf{Universal Category State Machine (UCSM)} for each category, which serves as a superset containing all potential functionalities within that category. Subsequently, we derived a \textbf{Specific State Machine (SSM)} for each individual object by pruning the UCSM based on the visual evidence of the object's components. For instance, if a specific weighing scale visually lacks a charging port, the corresponding charging states and transitions are removed from its SSM. This process yields a rigorous logic graph for every device, where transitions are bound to atomic actions and parameters. Component names are used only for internal state tracking and are not exposed to the model, ensuring the agent relies on visual features rather than textual shortcuts.

Notably, the generalization evaluated by VGEBench refers not to appearance-level novelty in static image recognition, but to \textit{functional and interactive generalization}: whether an agent can select actions and parameters based on visual components and revise incorrect hypotheses through environment feedback. Since the 3D assets are sourced from public websites, we cannot guarantee that closed-source VLMs have not encountered related assets or renderings during pretraining. However, the manually constructed Logic-Driven State Machines define instance-specific component-action bindings and state-transition rules that are not provided by the static visual assets. Thus, even if a model has seen images of similar devices, it cannot directly obtain the instance-specific interaction logic evaluated by our simulator from those images alone.

\subsection{Instruction Generation}
Based on the validated SSMs, we generated instruction-following tasks divided into single-turn and multi-turn scenarios. We performed pathfinding on the SSMs starting from a manually aligned initial state to generate state transition sequences that avoid loops. To simulate realistic user commands, we converted these symbolic sequences into natural language instructions using GPT-5-mini. In this process, the LLM is used only to verbalize a symbolic state-transition sequence whose initial state, target state, and valid transition path are determined by the validated SSM. For example, a symbolic transition \texttt{Off} $\rightarrow$ \texttt{Active} is refined into natural commands like ``Turn on the flashlight.'' In total, we constructed 4,948 single-turn and 10,005 multi-turn instructions. The detailed statistics of VGEBench and prompts used for instruction refinement can be found in the Appendix~\ref{sec:supplementary_statistics} and Tab.~\ref{tab:command_generation_prompt}.

To examine whether the evaluation is sensitive to the choice of instruction generator, we regenerate instructions using additional generators and evaluate the resulting variants. The results show that VGEBench is robust to the choice of generation model. Detailed results are provided in Appendix~\ref{sec:cross_model_instruction}.

\subsection{Environment Feedback and Task Evaluation}
\label{sec:env_feedback}
Inspired by ScienceWorld~\cite{wang-etal-2022-scienceworld}, we designed a closed-loop interactive system where the instantiated Specific State Machine (SSM) serves as the ground truth environment simulator. Unlike static evaluation datasets, our system evaluates the agent's ability to navigate through a state space by executing a sequence of visual actions. The interaction mechanism consists of two core components: the Feedback Generator and the Completion Judge.
\paragraph{Deterministic Feedback Generation}
At each time step $t$, the agent receives the current observation $O_t = (I_{view}, V_{list}, F_{t-1}, Inst_t)$, and executes an action $a_t = (\text{type}, \text{coord}, \text{params})$.
Here, $I_{\text{view}}$ represents the image of the current view, $V_{\text{list}}$ denotes the set of accessible views, $F_{t-1}$ is the feedback from the environment regarding the previous action, and $\text{Inst}_t$ denotes the current user instruction.
The environment simulator validates this action against the current state $S_t$ defined in the SSM. To simulate realistic physical constraints and provide instructive feedback, the system employs a hierarchical validation protocol:

\textbf{Validity Check:} The system first verifies if the action is permissible in the current view (e.g., physical interactions are prohibited in axonometric views like \textit{front\_top}, requiring a \texttt{Switch\_view} action first) and whether the coordinates fall within the image bounds. Invalid actions trigger immediate ``System Notification'' errors without altering the state.

\textbf{Geometric \& Semantic Matching:} If the action format is valid, the system queries the transitions defined in $S_t$. An action is considered successful only if:

\textbf{(1) $\text{coord}$:} The coordinates fall within the bounding box of a reactive component defined in the SSM for the current view.

\textbf{(2) $\text{action}$:} The action type matches the component's defined atomic action.

\textbf{(3) $\text{params}$:} The parameters satisfy the transition conditions (e.g., rotation direction).

\textbf{Environment Feedback:} To facilitate effective interaction and error recovery, the environment generates outcome-dependent feedback for each agent action. The feedback differentiates among qualitatively different interaction outcomes, enabling the agent to infer whether an action was successful, misapplied, or irrelevant.

\paragraph{Task Completion Criteria}
A task is defined not merely by reaching a final state, but by satisfying a sequence of sub-goals. Each task instance is associated with an ordered list of goal states $G = [g_1, g_2, \dots, g_n]$.
The agent can only observe the user instruction corresponding to the next goal after completing the current goal.
To limit resource consumption and prevent infinite exploration loops, we further control the interaction budget of the VLM using a Global Interaction Budget ($B_{\text{global}}$) and a Local Interaction Budget ($B_{\text{local}}$). Once the agent reaches either interaction limit, the interaction loop is immediately terminated, even if the task has not been fully completed.
Importantly, task completion is state-based rather than trajectory-based. Any valid sequence of interactions is accepted as long as it reaches the required goal states in the specified order within the interaction budget. Accordingly, when multiple components or action sequences can validly lead to the same goal according to the SSM, all such executions are accepted. The theoretical shortest path is used for efficiency-related evaluation rather than as the only valid execution trajectory.
The concrete feedback formulations and the interaction budget mechanism are deferred to Appendix~\ref{sec:appendix_environment}.

\subsection{Quality Control and Validation}
We implemented strict quality control throughout the dataset construction pipeline. At each stage, we randomly sampled 10\% of the annotated items for review, with each batch containing at least 1,000 items to ensure that our sampling strategy provides a statistically reliable estimate. An annotator’s annotations were accepted only if the error rate remained below 5\%; otherwise, their annotations were rejected and reassigned.
During Instruction Generation stage, inspired by MidMed\cite{shi-etal-2023-midmed}, we further filtered out unreasonable  state transition sequences to ensure that the resulting instructions remain natural, coherent, and executable.

After the dataset was constructed, following MT-bench~\cite{zheng2023mtbench}, we implemented a rigorous quality validation protocol on a subset randomly sampled to comprise 1\% of the entire dataset. We defined three metrics to evaluate data quality on a 3-level scale ($0, 1, 2$), which respectively assess the quality of vision, logic, and instruction alignment. Both human experts and two advanced VLMs (\textit{Gemini-3-Flash}, \textit{GPT-5-mini}) are employed for evaluation. As shown in Tab.~\ref{tab:qa_results}, our dataset achieves consistently high scores across all dimensions, with \textit{Gemini-3-flash/GPT-5-mini/Human Expert} reaching average scores of \textbf{1.978/1.947/1.943}. The strong human-model agreement further confirms the dataset's clarity and robustness. The complete quality-control workflow, data quality table and the scoring criteria are provided in Appendix~\ref{sec:validation_criteria}.

\subsection{Data statistics}

\begin{table}[t]
\centering
\small
\setlength{\tabcolsep}{3pt}
\renewcommand{\arraystretch}{1.1}
\begin{tabular}{ll @{\hspace{1em}} r}
\toprule
\textbf{Section} & \textbf{Item} & \textbf{Count} \\
\midrule
\multirow{3}{*}{\textbf{Assets}}
 & \# Categories & 26 \\
 & \# Devices & 968 \\
\midrule
\multirow{3}{*}{\textbf{Vision}}
 & \# Total views & 7,888 \\
 & \# Component types & 234 \\
 & \# Unique components & 3,712 \\
 & \# Annotated BBoxes & 7,264 \\
\midrule
\multirow{4}{*}{\textbf{Logic}}
 & \# UCSM & 26 \\
 & \hspace{1em}\textit{- States / Transitions} & 302 / 1,206 \\
 & \# SSM & 968 \\
 & \hspace{1em}\textit{- States / Transitions} & 6,352 / 15,861 \\
\midrule
\multirow{2}{*}{\textbf{Tasks}}
 & \# Total episodes & 14,953 \\
 & \hspace{1em}\textit{- Single / Multi-turn}  & 4,948 / 10,005 \\
\bottomrule
\end{tabular}
\caption{Data statistics of VGEBench. \textbf{UCSM}: Universal Category State Machine; \textbf{SSM}: Specific State Machine.}
\label{tab:statistics}
\end{table}

Tab.~\ref{tab:statistics} presents the data statistics of VGEBench.
We first curated a diverse collection of 968 high-precision 3D household device models spanning 26 categories.
To ensure rich interactability, we collected 7,888 views of these 3D models, recognized 3,712 unique interactive components in 234 classes (e.g., \textit{power buttons} or \textit{charging ports}), and annotated 7,264 BBoxes for these components.

To support realistic behavior Logic, we designed 26 Universal Category State Machines (UCSMs) as high-level template for each device category, comprising 302 states and 1,206 transitions. These were then instantiated into 968 Specific State Machines (SSMs) for individual devices to a total number of 6,352 states and 15,861 transitions. Based on this foundation, we generated \textbf{14,953} Task episodes. Notably, 10,005 of these are multi-turn interactions, which imposes significant challenges for long-horizon planning and exploration.

\section{Experiment}

\begin{table*}[t!]
  \centering
  \resizebox{\linewidth}{!}{
  \begin{tabular}{lccccccccc}
    \toprule
    \multirow{2}{*}{Model} &
    \multicolumn{2}{c}{Task Performance} &
    \multicolumn{2}{c}{Efficiency} &
    \multicolumn{4}{c}{Visual Grounding and Perception} &
    \multicolumn{1}{c}{Exploration} \\
    \cmidrule(lr){2-3}\cmidrule(lr){4-5}\cmidrule(lr){6-9}\cmidrule(lr){10-10}
    & SR $\uparrow$ & SSR $\uparrow$ & SPL $\uparrow$ & State-F1 $\uparrow$
    & EIR $\uparrow$ & TIR $\uparrow$ & VSPS $\downarrow$ & GVPE $\downarrow$
    & EER $\uparrow$ \\
    \midrule
    MiMo-Embodied-7B
    & 1.48 & 2.79 & 1.10 & 47.86 & 7.34 & 3.44 & \textbf{0.88} & 21.66 & 15.78 \\
    InternVL3.5-8B-Instruct
      & 3.72 & 7.71 & 2.28 & 51.07 & 23.14 & 13.12 & 1.23 & 11.26 & 20.89 \\
    Qwen3-VL-8B-Instruct
      & 7.77 & 12.69 & 5.11 & 54.31 & 30.46 & 14.80 & 1.42 & 11.46 & 29.91 \\
    GPT-5-mini
      & 10.78 & 17.51 & 5.88 & 57.16 & 22.86 & 10.21 & 1.20 & 6.72 & 28.81 \\
    Doubao-1.5-Thinking-Vision-Pro
      & \underline{16.68} & \underline{24.30} & \underline{12.50} & \underline{60.61} & \underline{46.62} & \underline{23.93} & 1.21 & \underline{3.21} & \underline{37.24} \\
    Gemini-3-Flash
      & \textbf{54.27} & \textbf{62.86} & \textbf{39.34} & \textbf{78.67} & \textbf{67.21} & \textbf{43.56} & \underline{1.12} & \textbf{1.85} & \textbf{64.36} \\
    \bottomrule
  \end{tabular}
  }
  \caption{\textbf{Main results on VGEBench.} \textbf{Bold} indicates the best results, and \underline{underline} indicates the second-best results. $\uparrow$ indicates higher is better, and $\downarrow$ indicates lower is better.}
  \label{tab:main_results}
\end{table*}

\subsection{Experimental Setups}

\paragraph{Models.} We conduct a comprehensive experiment on \textbf{VGEBench} across three categories of VLMs: (a) Closed-source VLMs: GPT-5-mini~\cite{openai2025gpt5}, Gemini-3-Flash~\cite{google2025gemini3}, Doubao-1.5-Thinking-Vision-Pro~\cite{guo2025seed15vltechnicalreport}; (b) Open-source VLMs: Qwen3-VL-8B-Instruct~\cite{bai2025qwen3vltechnicalreport}, InternVL3.5-8B~\cite{wang2025internvl35advancingopensourcemultimodal}, Mimo-Embodied-7B~\cite{hao2025mimoembodiedxembodiedfoundationmodel}.

\paragraph{Implementation.} VGEBench is a test-only benchmark and does not define conventional train/test splits. For all models, we adopt ReAct~\cite{yao2023react} as the underlying reasoning framework and expose the complete chat history. For interaction budget control, the redundancy factor $\lambda$ is set to $5$. More details about the reasoning framework and the interaction budget can be found in Appendices~\ref{sec:method_prompt} and~\ref{sec:appendix_budget}.

\subsection{Metrics}
\label{sec:metrics}

Inspired by classic works in embodied~\cite{anderson2018evaluation} and vision~\cite{yu2016modeling} domains, we adopt a multi-dimensional assessment protocol covering task performance, execution efficiency, visual grounding, and reasoning stability to evaluate the capabilities of VLMs comprehensively. Detailed definitions and formulations are provided in Appendix~\ref{sec:metrics_details}.

\subsubsection{Task Performance}
We primarily measure the \textbf{Success Rate (SR)} to denote the percentage of successfully completed episodes. To more accurately evaluate multi-turn tasks, we additionally report the \textbf{Sub-task Success Rate (SSR)} to assess step-wise correctness based on achieved sub-goals.

\subsubsection{Efficiency}
We report \textbf{Success Weighted by Path Length (SPL)} to jointly account for success rate and execution length. We also compute the \textbf{State-F1 Score} to measure the alignment between the model’s state trajectory and the ground-truth optimal path.

\subsubsection{Visual Grounding and Navigation}
We assess visual interaction using the \textbf{Effective Interaction Rate (EIR)} (validity of component identification) and the \textbf{Target Interaction Rate (TIR)} (precision of goal-oriented grounding). Navigation efficiency is evaluated via \textbf{View Switches Per Success (VSPS)} for local efficiency and \textbf{Global View Switches Per Episode (GVPE)} for macro-level exploration cost.

\subsubsection{Exploration}
We define the \textbf{Effective Exploration Rate (EER)} as the ratio of valid operations to total steps, serving as an indicator of reasoning efficiency and safety awareness.

\subsection{Main Results}
Tab.~\ref{tab:main_results} presents the comprehensive evaluation of representative VLMs on VGEBench. We report the results across nine metrics described in Section~\ref{sec:metrics}.

\paragraph{VGEBench poses a significant challenge to VLM-based agents.}

The result shows that only Gemini-3-Flash achieves a notable Success Rate of 54.27\%, while the second-best model, Doubao-1.5-Thinking-Vision-Pro, drops sharply to 16.68\%. All remaining models obtain success rates below 15\%. Even when considering subtask completion, no model other than Gemini-3-Flash attains an SSR exceeding 25\%. These results indicate that generalizable visually grounded exploration remains a difficult open problem for the majority of VLMs.

\paragraph{Both visual grounding and exploration capabilities determine model performance.} 
The results demonstrate that advanced VLMs, especially \textbf{Gemini-3-Flash} and \textbf{Doubao-1.5-Thinking-Vision-Pro}, achieve significantly higher scores in both visual grounding (EIR, TIR) and exploration (EER) compared to other models. This superior capability to accurately perceive targets and efficiently explore the environment directly underpins their leading task completion rates. In contrast, \textbf{MiMo-Embodied-7B} presents a compelling counter-example. Benefiting from its specialized training in autonomous driving scenarios, it exhibits exceptional strength in View Navigation (VSPS), achieving a score of 0.88 which even surpasses Gemini-3-Flash. However, its overall task performance remains poor (SR: 1.48\%) due to a critical lack of fine-grained visual grounding and reasoning capabilities. This disparity highlights that while navigation efficiency is valuable, the integration of fine-grained visual perception and logical reasoning is the decisive factor for success in complex household environments.

\section{Analysis}

To better evaluate the ability of VLMs in exploring household devices and further investigate the underlying mechanisms, we propose four research questions (RQs). These RQs include four key dimensions in VGEBench: (1) the necessity of generalization capacity, (2) the improvement from extended interaction budgets, (3) the primary visual bottleneck, and (4) the gap between current VLMs and human performance.

\subsection*{RQ1: The necessity of generalization capacity: Can VLMs immediately recognize target?}

\begin{table}[th]
    \centering
    \resizebox{\linewidth}{!}{
        \begin{tabular}{lcc}
            \toprule
            \textbf{Model} & \textbf{PSR} & \textbf{SSR} \\
            \midrule
            MiMo-Embodied-7B & 0.51\% & 2.79\% \\
            Qwen3-VL-8B & 1.53\% & 12.69\% \\
            InternVL3.5-8B & 0.91\% & 7.71\% \\
            GPT-5-mini & 1.97\% & 17.51\% \\
            Doubao-1.5-Thinking-Vision-Pro & 4.06\% & 24.30\% \\
            Gemini-3-Flash & 12.94\% & 62.86\% \\
            \bottomrule
        \end{tabular}
    }
    \caption{\textbf{Generalization Performance.} We report \textit{SSR} and \textit{PSR} (Perfect Sub-task success Rate): The proportion of sub-tasks completed using exactly the optimal number of interaction steps.}
    \label{tab:generalize}
\end{table}

To investigate the necessity of generalization capacity in VLMs, we evaluate whether models can achieve ``one-shot success'' in Table~\ref{tab:generalize}. The results reveal a pronounced discrepancy between PSR and SSR across all models. Even for the best-performing model, Gemini-3-Flash, the PSR is only 12.94\%, whereas the SSR reaches 62.86\%. This gap indicates that current models exhibit little to no immediate, zero-shot understanding of the operational logic of household devices. Instead, successful performance mainly arises from feedback-driven exploration, where agents iteratively test hypotheses, correct errors, and gradually infer device-specific functional logic.

\subsection*{RQ2: Do VLMs continuously benefit from extended exploration budgets?}

\begin{figure}[th]
  \centering
  \includegraphics[width=\linewidth]{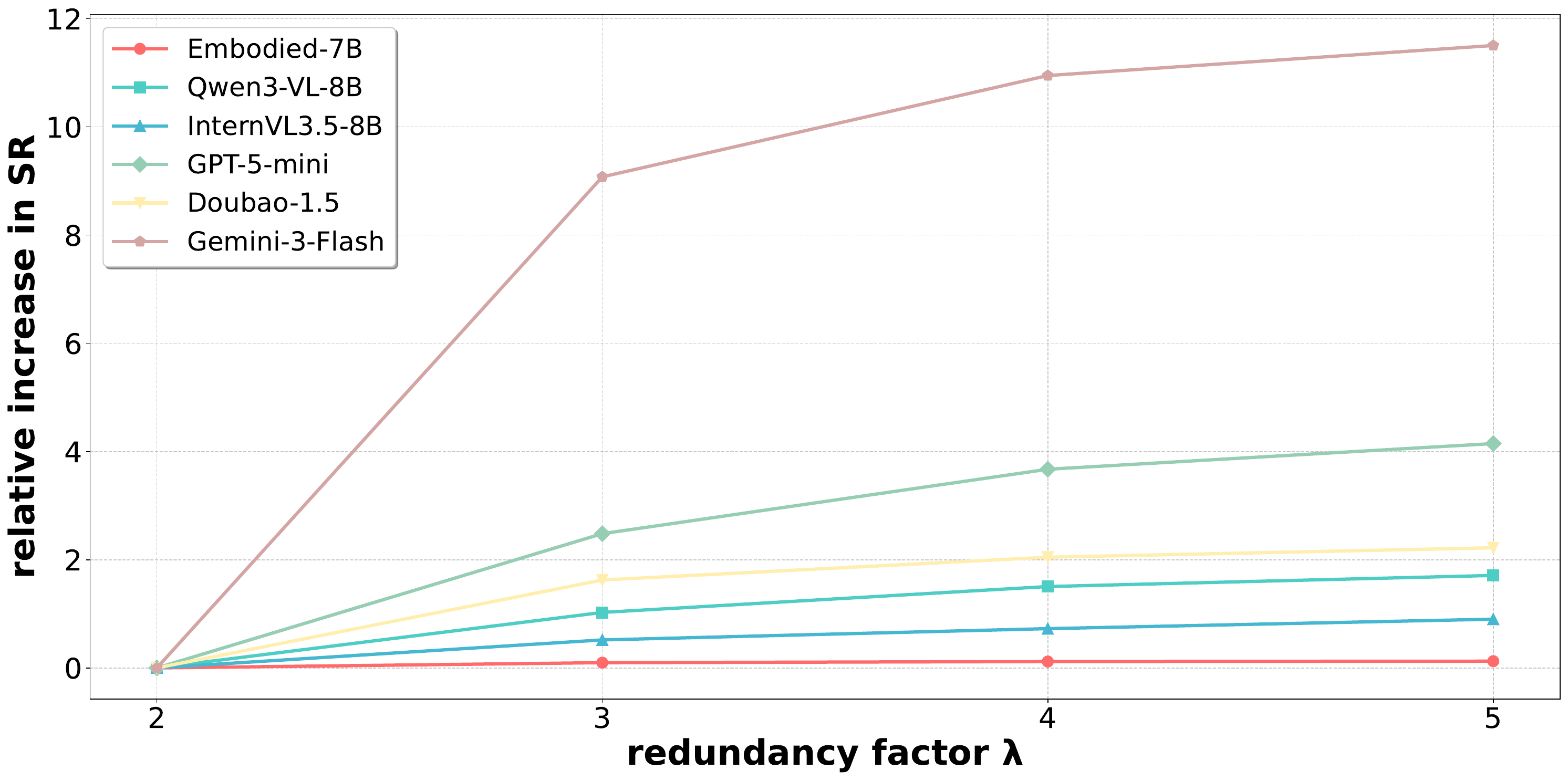}
  \caption{\textbf{Impact of interaction budget on SR score.} We normalize the performance at the baseline budget ($\lambda = 2$) to zero to isolate the relative gains. The curves illustrate the relative improvement as $\lambda$ increases.}
  \label{fig:interaction_comparison}
\end{figure}

Fig.~\ref{fig:interaction_comparison} illustrates the relative performance gains as the interaction limit increases (the detailed relationship between $\lambda$ and the interaction limit can be found in Appendix~\ref{sec:appendix_budget}). All models exhibit a pattern of improvement that gradually tapers off, indicating that current VLMs are able to convert additional interactions into better task completion, especially during the early stages of exploration. However, weaker models (e.g., Qwen3-VL-8B) show a more pronounced and earlier performance plateau, whereas advanced models such as Gemini-3-Flash, benefiting from stronger exploratory reasoning and more stable long-term memory maintenance, demonstrate a more robust ability to exploit deep exploration for effective self-correction. In addition to the interaction budget, we further examine how the available interaction history affects agents' self-correction ability. The detailed analysis is provided in Appendix~\ref{sec:history-sensitivity}.

\subsection*{RQ3: Where lies the visual bottleneck: Coarse-grained navigation or fine-grained grounding?}

\begin{figure}[th]
  \centering
  \includegraphics[width=\linewidth]{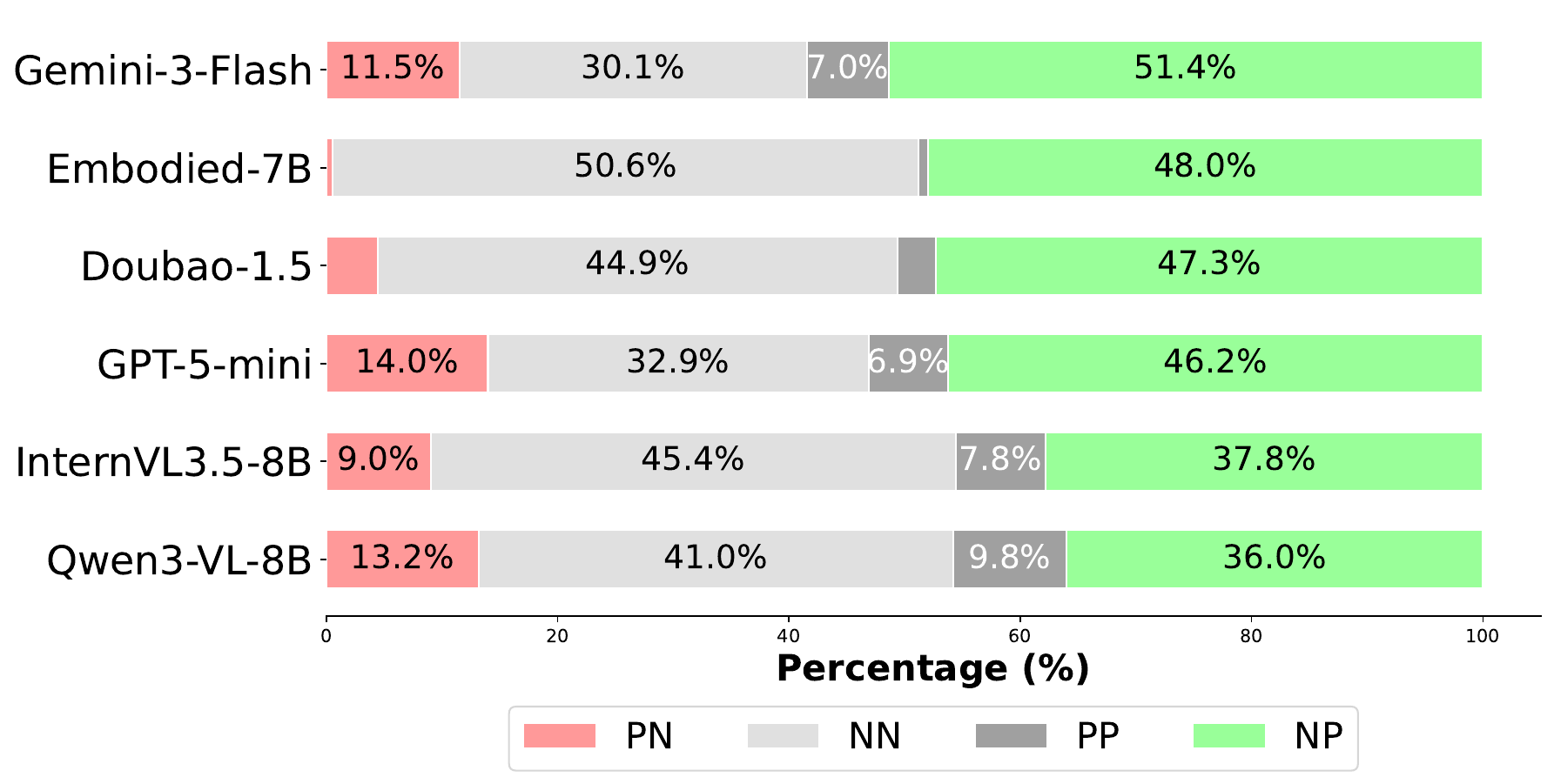}
  \caption{\textbf{Distribution of view switching actions.} We classify \texttt{Switch\_view} actions into four categories based on whether the target component is visible in the start and end views (defined as "Positive" or "Negative"). \textbf{NP (Green)}: Successful navigation to the target view (Negative $\to$ Positive view); \textbf{PN (Red)}: Losing the target; \textbf{NN}: Ineffective searching; \textbf{PP}: Maintaining focus.}
  \label{fig:switch_view_comparison}
\end{figure}

\begin{table}[t]
  \centering
  \resizebox{\linewidth}{!}{
  \begin{tabular}{lcc}
    \toprule
    \textbf{Model} & \textbf{Avg Dist} $\downarrow$ & \textbf{Avg W-Dist} $\downarrow$ \\
    \midrule
    Mimo-Embodied-7B & 274.04 & 441.69 \\
    InternVL3.5-8B-Instruct & 188.35 & 312.43 \\
    Qwen3-VL-8B & 178.09 & 312.63 \\
    Doubao-1.5-Thinking-Vision-Pro & 188.53 & 290.53 \\
    GPT-5-mini & 178.96 & 290.33 \\
    Gemini-3-Flash & \textbf{109.18} & \textbf{163.75} \\
    \bottomrule
  \end{tabular}
  }
  \caption{\textbf{Fine-grained Action Precision Analysis.} Metrics are calculated on a normalized $1000 \times 1000$ scale. \textit{Avg Dist}: The average distance to the closest target BBox. \textit{Avg W-Dist}: The weighted average distance that penalizes repeated failed attempts on the same target.}
  \label{tab:action_distance}
\end{table}

To isolate the bottleneck, we compared performance in coarse-grained navigation (Fig.~\ref{fig:switch_view_comparison}) with fine-grained grounding (Tab.~\ref{tab:action_distance}). The performance gap in navigation is relatively moderate: For example, GPT-5-mini's successful navigation (\textit{NP}) is only 12.5\% lower than that of Gemini-3-Flash. However, the disparity in fine-grained grounding is drastic. GPT-5-mini's coordinate error (\textit{Avg Dist}) is 63.9\% worse than Gemini-3-Flash. Crucially, the even wider gap in \textit{Avg W-Dist} (77.3\%) reveals that weaker models fail to efficiently correct their coordinates after initial failures. This empirical result confirms that the primary bottleneck in VGEBench is not navigating to the correct view, but precisely localizing the interactive component within it. To further examine the fine-grained grounding bottleneck, we additionally evaluate model robustness under common visual perturbations. The detailed results are reported in Appendix~\ref{sec:visual-perturbation}.

\subsection*{RQ4: How large is the gap between humans and current VLMs?}

To contextualize current VLM performance, we evaluate a human baseline on 149 episodes (approximately 1\% of VGEBench) under the same interface and interaction budgets as Gemini-3-Flash. As shown in Tab.~\ref{tab:human_main}, humans achieve consistently higher task performance, reaching 74.50\% SR and 81.43\% SSR, compared with 56.38\% and 64.32\% for Gemini-3-Flash on the same episodes. Human performance nevertheless remains below saturation, indicating that manual-free operation of unfamiliar devices remains non-trivial even for humans under the constrained interaction setting.

\begin{table}[t]
    \centering
    \small
    \begin{tabular}{lccc}
        \toprule
        Model & SR & SSR & SPL\\
        \midrule
        Gemini-3-Flash & 56.38 & 64.32 & 39.23\\
        Human Baseline & \textbf{74.50} & \textbf{81.43} & \textbf{54.86} \\
        \bottomrule
    \end{tabular}
    \caption{Comparison between Gemini-3-Flash and Human baseline on 149 episodes under the same interface and interaction budgets.}
    \label{tab:human_main}
\end{table}

Beyond task success, the complete metric and error analyses reveal a notable behavioral difference: humans produce fewer Invalid Coord Errors and therefore more frequently interact with valid component regions, enabling more informative component-level feedback for subsequent hypothesis refinement. This provides additional evidence that fine-grained visual grounding is important for efficient exploration. Full results and error-distribution comparisons are provided in Appendix~\ref{sec:human_baseline}.

\section{Conclusion}

To investigate the capabilities of current VLMs in visually grounded device exploration, this paper introduces VGEBench, a comprehensive benchmark designed to evaluate generalizable visually grounded exploration without reliance on manuals. We construct a Logic-Driven State Machine framework which simulates multi-turn interaction loops to provide deterministic feedback. Extensive experiments reveal that despite strong semantic knowledge, current VLMs still struggle significantly with fine-grained visual grounding and maintaining long-horizon state consistency. We hope that VGEBench will inspire further research to bridge the gap between abstract world knowledge and precise device execution.

\section*{Limitations}

VGEBench introduces a dynamic, interactive evaluation framework involving multi-turn agent-environment feedback loops. Consequently, conducting a full-scale evaluation requires significant computational resources and inference time. Additionally, the necessity for high-resolution visual observations to resolve fine-grained grounding details imposes high demands on the visual encoding efficiency and context window of current VLMs.

VGEBench employs interactive components annotated on high-fidelity rendered views, discrete atomic actions, and fixed viewpoint switching. This abstraction of low-level control to focus on high-level reasoning avoids the confounding effects of factors such as continuous robotic control and free-form camera navigation, thereby enabling a more focused evaluation of models' capabilities in the generalizable visually grounded exploration task. Although this design enables deterministic evaluation, it does not fully capture all sim-to-real challenges encountered in physical robotic deployment.

The current observations are device-centered rendered images and do not fully capture complex backgrounds, natural illumination, or object occlusion in real environments. Although Appendix~\ref{sec:visual-perturbation} provides controlled robustness tests under compression, blur, and random occlusion, these perturbations cannot fully substitute for real-world observations.

\section*{Ethics Statement}

We are committed to strict ethical standards in the construction and distribution of VGEBench.

\paragraph{Data Compliance.}
Our dataset leverages 3D assets sourced from open platforms (Sketchfab and 3D Warehouse). We strictly adhered to the platforms' ethical and licensing guidelines. Specifically, we utilized only assets available under permissible licenses and explicitly excluded any assets marked with ``NoAI'' tags to respect creators' rights regarding generative AI training. The dataset focuses solely on household objects and contains no Personally Identifiable Information (PII) or offensive content.

\paragraph{Professional Annotation.}
We ensure that all employed human annotators underwent rigorous training specific to the categories of household devices involved in the dataset. All participants were fully informed of the intended data usage prior to their involvement. Additionally, we have provided fair and reasonable compensation for their work, ensuring that their efforts are appropriately rewarded and that the quality of the annotated content is guaranteed.

\section*{Acknowledgments}

We thank all the reviewers for their insightful and valuable comments. This work is supported by the National Natural Science Foundation
of China (Grant No.U21B2009).

\bibliography{custom}

\clearpage
\appendix

\section*{Appendix}
\label{sec:appendix}

\section{Supplementary Statistics}
\label{sec:supplementary_statistics}

This section reports supplementary statistics and explains the criteria used for selecting device categories in VGEBench.

\begin{table}[h]
\centering
\begin{tabular}{lccc}
\toprule
Category & Total & SF & 3DW \\
\midrule
CoffeeMachine & 72 & 52 & 20 \\
DigitalAlarmClock & 71 & 66 & 5 \\
Lighter & 63 & 46 & 17 \\
Microwave & 59 & 48 & 11 \\
WashingMachine & 55 & 50 & 5 \\
MechanicalAlarmClock & 54 & 51 & 3 \\
WeighingScale & 49 & 44 & 5 \\
Heater & 48 & 37 & 11 \\
Lamp & 46 & 45 & 1 \\
ElectricKettle & 44 & 39 & 5 \\
ElectricFan & 41 & 32 & 9 \\
AirFryer & 36 & 27 & 9 \\
Flashlight & 34 & 22 & 12 \\
FilmCamera & 32 & 24 & 8 \\
Razor & 29 & 27 & 2 \\
DigitalCamera & 28 & 24 & 4 \\
HairDryer & 27 & 21 & 6 \\
Humidifier & 27 & 23 & 4 \\
Landline & 23 & 19 & 4 \\
PowerBank & 23 & 19 & 4 \\
CarKey & 22 & 14 & 8 \\
Vacuum & 21 & 19 & 2 \\
AirPurifier & 20 & 18 & 2 \\
RiceCooker & 19 & 13 & 6 \\
ElectricToothbrush & 14 & 13 & 1 \\
ThermometerGun & 11 & 11 & 0 \\
\midrule
\textbf{Total} & \textbf{968} & \textbf{804} & \textbf{164} \\
\bottomrule
\end{tabular}
\caption{Category-wise distribution of 3D models. (SF: Sketchfab; 3DW: 3DWarehouse)}
\label{tab:category_distribution}
\end{table}

\paragraph{Category Selection Criteria.}
We prioritize household devices that are commonly encountered in everyday life, such as \textit{CoffeeMachine}, \textit{Microwave}, \textit{Lamp}, and \textit{DigitalAlarmClock}.
This choice ensures that evaluated models can rely on world knowledge rather than domain-specific expertise.
At the same time, we restrict the selection to devices with relatively simple and compact physical structures, avoiding objects with deeply nested or highly articulated components, so as to prevent incomplete visual coverage from becoming a confounding factor in evaluation.

\begin{table*}[t]
  \centering
  \begin{tabular}{lrrrrrrr}
    \toprule
    \textbf{Category} & \textbf{press} & \textbf{pull} & \textbf{push} & \textbf{rotate} & \textbf{grasp} & \textbf{timed\_wait} & \textbf{Total} \\
    \midrule
    AirFryer & 813 & 44 & 1 & 1051 & 0 & 0 & 1909 \\
    AirPurifier & 268 & 0 & 0 & 6 & 0 & 0 & 274 \\
    CarKey & 147 & 0 & 0 & 0 & 0 & 0 & 147 \\
    CoffeeMachine & 752 & 86 & 55 & 63 & 10 & 366 & 1332 \\
    DigitalAlarmClock & 314 & 11 & 118 & 42 & 0 & 45 & 530 \\
    DigitalCamera & 92 & 19 & 34 & 99 & 0 & 0 & 244 \\
    ElectricFan & 809 & 111 & 103 & 242 & 0 & 0 & 1265 \\
    ElectricKettle & 85 & 4 & 167 & 9 & 0 & 43 & 308 \\
    ElectricToothbrush & 66 & 6 & 34 & 56 & 14 & 0 & 176 \\
    FilmCamera & 77 & 0 & 8 & 245 & 0 & 0 & 330 \\
    Flashlight & 90 & 23 & 7 & 141 & 0 & 0 & 261 \\
    HairDryer & 298 & 0 & 672 & 0 & 296 & 0 & 1266 \\
    Heater & 177 & 7 & 2 & 270 & 0 & 0 & 456 \\
    Humidifier & 194 & 0 & 0 & 8 & 0 & 0 & 202 \\
    Lamp & 92 & 0 & 0 & 0 & 0 & 0 & 92 \\
    Landline & 66 & 0 & 0 & 10 & 65 & 0 & 141 \\
    Lighter & 124 & 0 & 92 & 165 & 0 & 49 & 430 \\
    MechanicalAlarmClock & 47 & 9 & 57 & 182 & 0 & 16 & 311 \\
    Microwave & 818 & 39 & 0 & 1304 & 0 & 0 & 2161 \\
    PowerBank & 56 & 45 & 45 & 0 & 0 & 0 & 146 \\
    Razor & 76 & 58 & 66 & 0 & 0 & 0 & 200 \\
    RiceCooker & 174 & 56 & 2 & 0 & 0 & 9 & 241 \\
    ThermometerGun & 21 & 2 & 2 & 0 & 0 & 0 & 25 \\
    Vacuum & 43 & 10 & 51 & 42 & 0 & 0 & 146 \\
    WashingMachine & 508 & 345 & 265 & 91 & 0 & 62 & 1271 \\
    WeighingScale & 304 & 14 & 210 & 12 & 0 & 0 & 540 \\
    \midrule
    \textbf{Total} & \textbf{6511} & \textbf{889} & \textbf{1991} & \textbf{4038} & \textbf{385} & \textbf{590} & \textbf{14404} \\
    \bottomrule
  \end{tabular}
  \caption{Statistical distribution of atomic actions across 26 device categories.}
  \label{tab:action_distribution}
\end{table*}

\paragraph{Data Source Distribution.}
As shown in Tab.~\ref{tab:category_distribution}, the benchmark contains 968 tool instances spanning 26 categories, sourced from both Sketchfab and 3DWarehouse. In our data sources, Sketchfab accounts for a higher proportion due to its greater availability of high-quality 3D models. On the other hand, 3DWarehouse, as a completely free 3D model website, has a relatively smaller number of high-quality models.

\paragraph{Action Type Distribution.}

Tab.~\ref{tab:action_distribution} presents the statistical distribution of atomic actions in SSM transitions across 26 device categories. \texttt{Press} serves as the most fundamental action, appearing ubiquitously across all device types and accounting for approximately 45.20\% of the total dataset. In contrast, \texttt{Rotate}, \texttt{Grasp} and \texttt{Timed\_wait} exhibit significant device-dependency; they are sparse across the dataset and appear only in specific categories compatible with such affordances. Furthermore, \texttt{Pull} and \texttt{Push} demonstrate a notable correlation in their occurrence frequencies within many devices (e.g., \textit{ElectricFan}, \textit{Powerbank}), as these actions typically function as complementary pairs in mechanical operations, such as opening and closing drawers.

\section{Error Analysis}
\label{sec:error_analysis}

\begin{figure}[h] 
\centering 
\includegraphics[width=\linewidth]{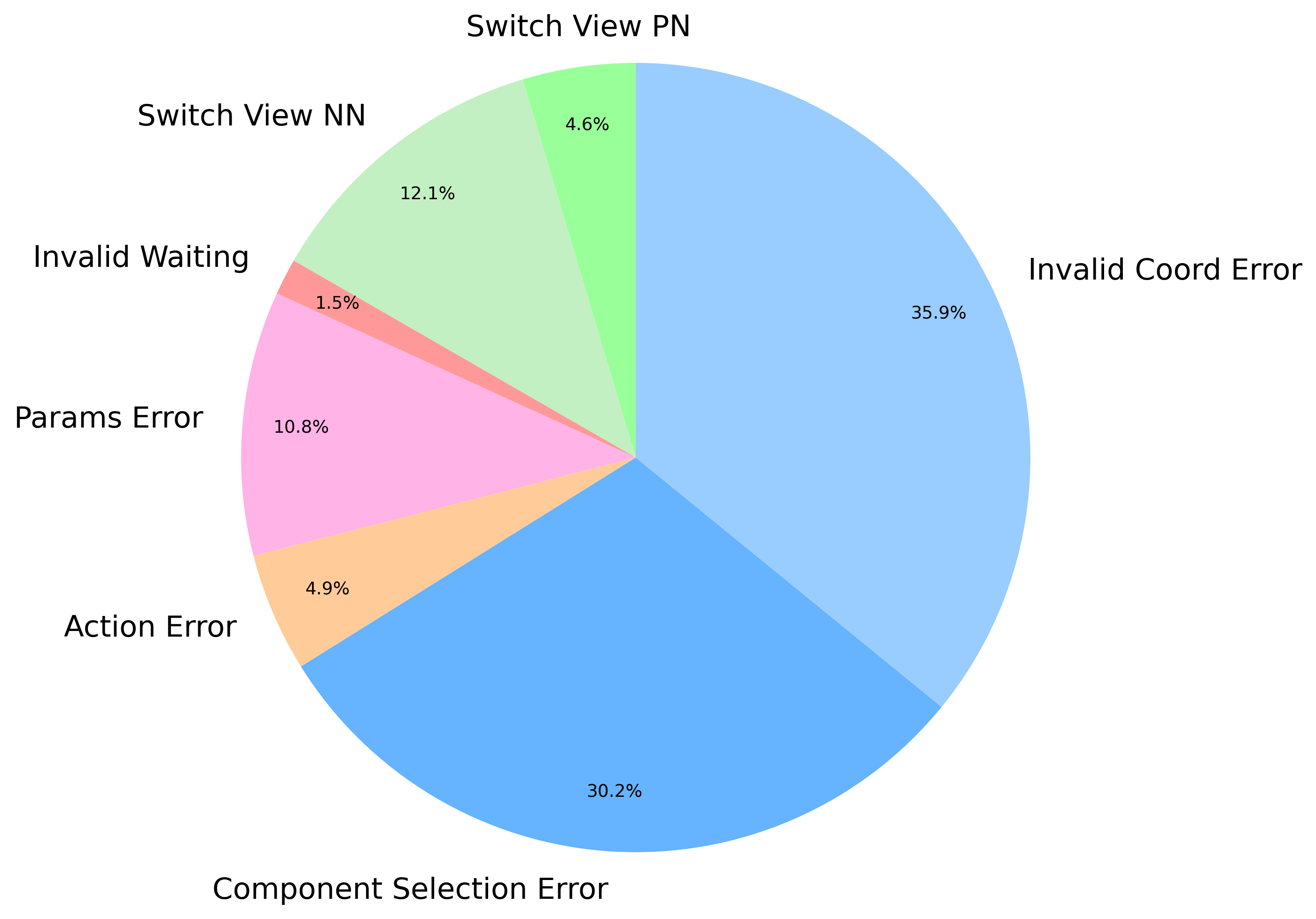}
\caption{Distribution of error types. The errors are grouped into Fine-grained Visual Grounding (blue), Action Execution (red), and Visual Navigation (green) categories.}
\label{fig:error_pie}
\end{figure}

To deeply understand the limitations of current advanced VLMs in performing generalizable visually grounded exploration, we conducted a systematic error analysis on the failure cases collected during the evaluation of Gemini-3-Flash, which serves as the best-performing model in our benchmark. We categorize the observed errors into three distinct groups: Fine-grained Visual Grounding, Action Execution, and Visual Navigation. These error categories are defined as follows: 

\textbf{(1)Fine-grained Visual Grounding Errors:} Failures in pixel-level localization, including \textit{Invalid Coord Error} (clicking on non-interactive regions) and \textit{Component Selection Error} (interacting with incorrect or non-optimal components). 

\textbf{(2)Action Execution Errors:} Correct targeting but incorrect operation logic, comprising \textit{Params Error} (wrong execution parameters, e.g., rotation direction), \textit{Action Error} (mismatched atomic actions), and \textit{Invalid Waiting} (waiting without triggering state changes). 

\textbf{(3)Visual Navigation Errors:} Ineffective view exploration, specifically \textit{Switch View NN} (ineffective searching) and \textit{Switch View PN} (losing the target component). 

Statistical analysis reveals that fine-grained visual grounding is the dominant failure mode, accounting for approximately 66.1\% of all errors. This is significantly higher than Action Execution (17.2\%) and Visual Navigation (16.7\%). The high prevalence of coordinate-related errors indicates that precise spatial localization remains the primary bottleneck for current VLMs within the device-centric interaction setting of VGEBench.

In conclusion, our analysis highlights a dichotomy in model performance: while agents demonstrate competent global navigation and high-level planning, they exhibit significant weakness in \textbf{fine-grained visual perception}. Regarding reasoning, the high rate of parameter errors (10.8\%) indicates a struggle with precise physical constraints even when the correct component is identified. Typical examples and the proportions for each category are shown in Tab.~\ref{tab:error_analysis} and Fig.~\ref{fig:error_pie}.

\begin{table*}[htbp]
    \centering
    \begin{tabular}{l c >{\centering\arraybackslash}m{8cm}}
        \toprule
        \textbf{Error Type} & \textbf{Current View} & \textbf{Model Output} \\
        \midrule
        
        Invalid Coord Error & 
        \includegraphics[max width=3cm, max height=3cm, keepaspectratio, valign=m]{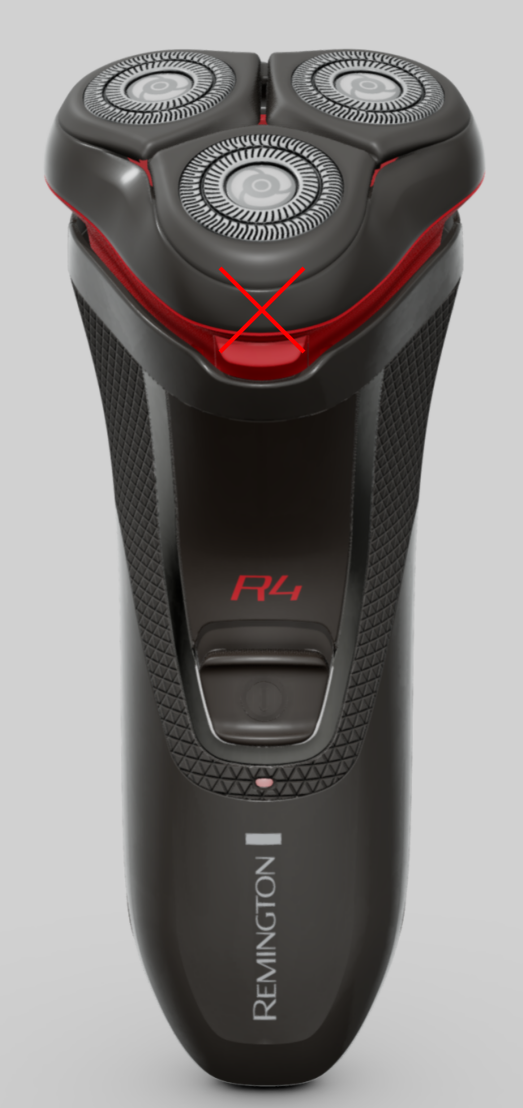} & 
        \textbf{Thinking}: To detach the blade head from the shaver, I need to locate the release mechanism. In the front view, there is a prominent red button located just below the blade head assembly. Pressing this button should release the latch holding the blade head.

        \textbf{Action}: Press(coord=[500, 280], params=[0]) \\
        \midrule
        
        \shortstack[l]{Component \\ Selection Error} & 
        \includegraphics[max width=3cm, max height=3cm, keepaspectratio, valign=m]{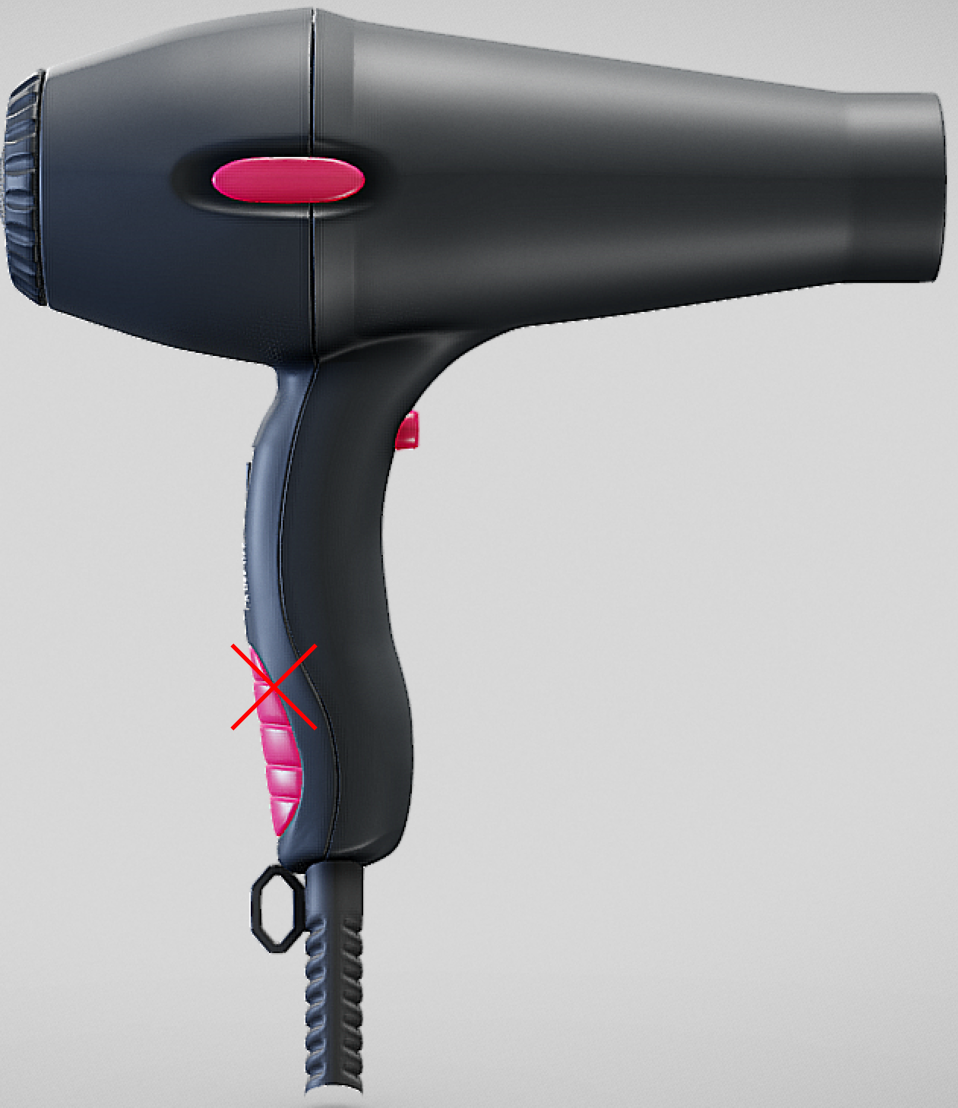} & 
        \textbf{Thinking}: To set the dryer to hot air mode, I need to adjust the temperature switch, which is the upper sliding switch on the handle. The lower switch controls the fan speed and should remain in the 'off' position as per the instruction. I will push the upper switch to its highest position.

        \textbf{Action}: Push(coord=[285, 620], params=[0]) \\
        \midrule
        
        Params Error & 
        \includegraphics[max width=3cm, max height=3cm, keepaspectratio, valign=m]{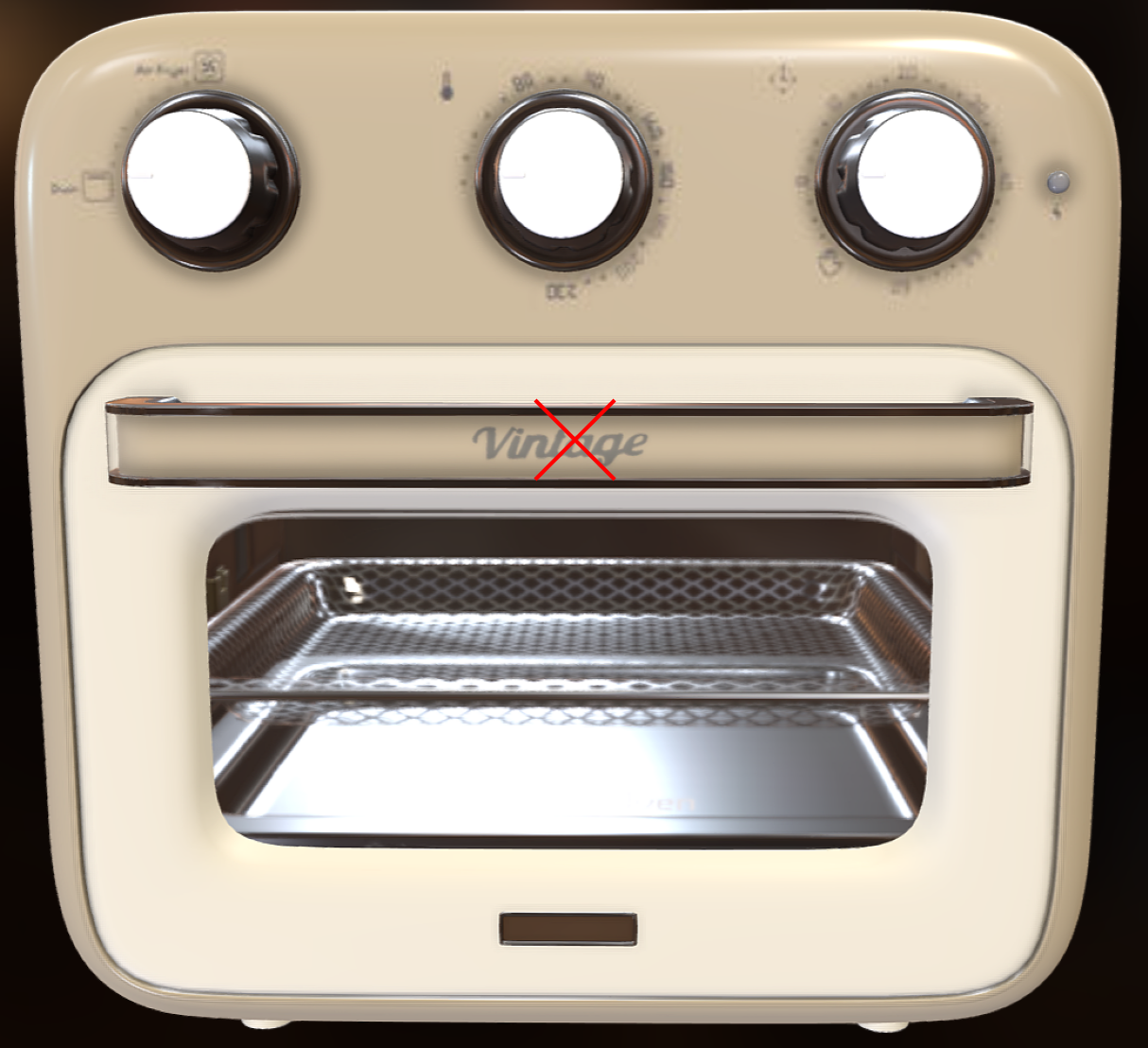} &
        \textbf{Thinking}: To load the food, I first need to open the air fryer door. I will pull the handle located in the center of the door downwards.
        
        \textbf{Action}: Pull(coord=[500, 420], params=[1])\\
        \midrule
        
        Action Error & 
        \includegraphics[max width=3cm, max height=3cm, keepaspectratio, valign=m]{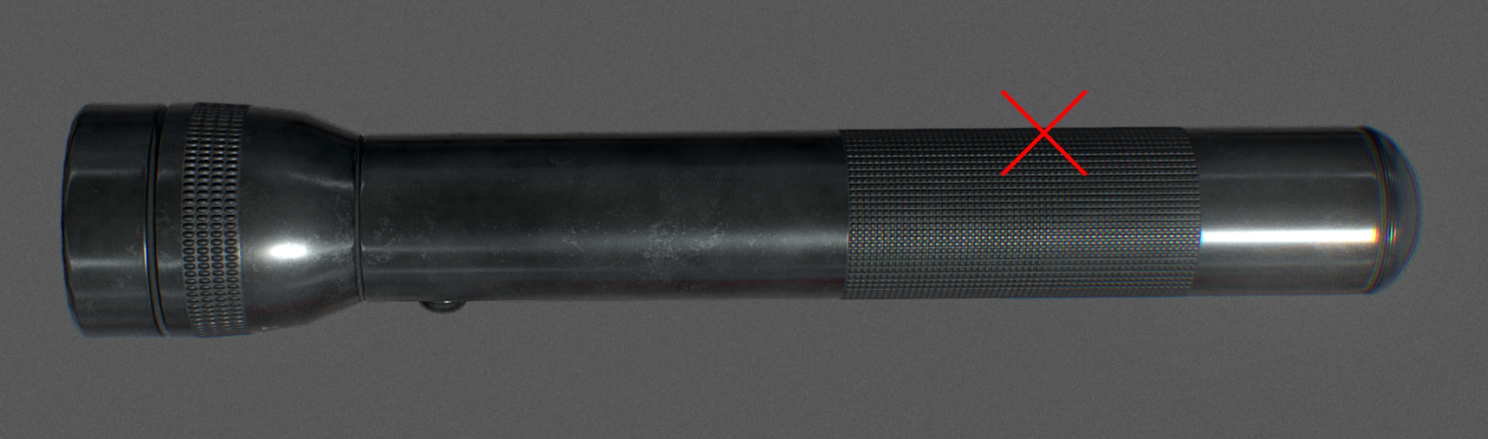} & 
        \textbf{Thinking}: The flashlight has a button visible in the current `top` view. It's located on the neck of the flashlight. I will click on it to turn it on.

        \textbf{Action}: Click(params=[701, 303]) \\
        \midrule
        
        Invalid Waiting & 
        \includegraphics[max width=3cm, max height=3cm, keepaspectratio, valign=m]{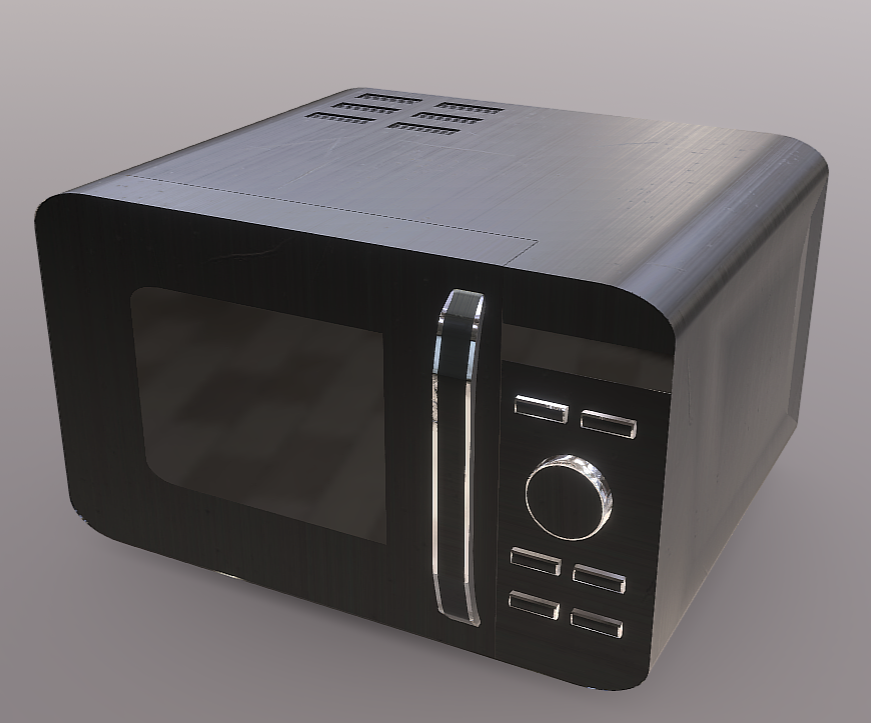} & 
        \textbf{Thinking}: The microwave is currently operating at medium power in short-time mode with the food inside, which fulfills the user's instruction. I will wait for the cooking process to complete.

        \textbf{Action}: Timed\_wait() \\
        \midrule
        
        Switch View NN & 
        \begin{tabular}{@{}c@{}}
            \includegraphics[max width=1.5cm, keepaspectratio, valign=m]{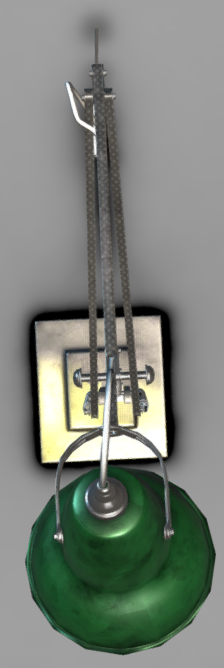} $\to$ \includegraphics[max width=1.5cm, keepaspectratio, valign=m]{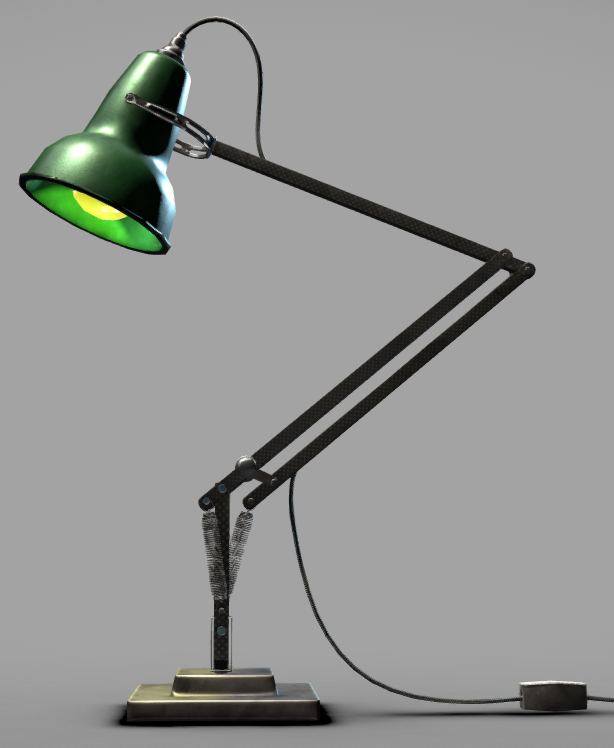}
        \end{tabular} & 
        \textbf{Thinking}: The user wants to turn on the desk lamp. In the top view, I can see the lamp's structure, but the switch on the cord is not clearly visible. I will switch to the right view to get a better look at the base and the cord, which might reveal the switch more clearly.

        \textbf{Action}: Switch\_view(params=[right]) \\
        \midrule
        
        Switch View PN & 
        \begin{tabular}{@{}c@{}}
            \includegraphics[max width=1.5cm, keepaspectratio, valign=m]{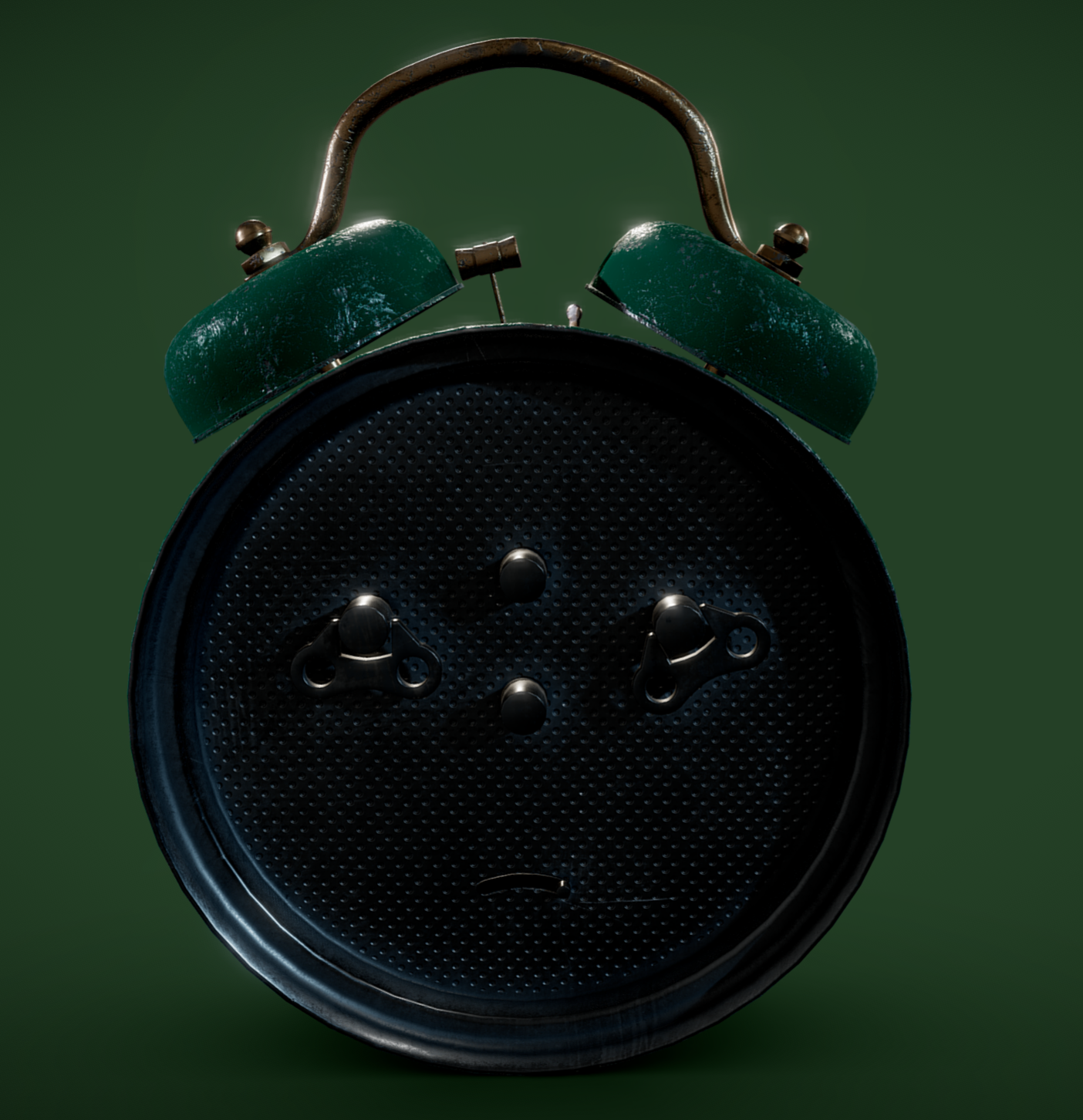} $\to$ \includegraphics[max width=1.5cm, keepaspectratio, valign=m]{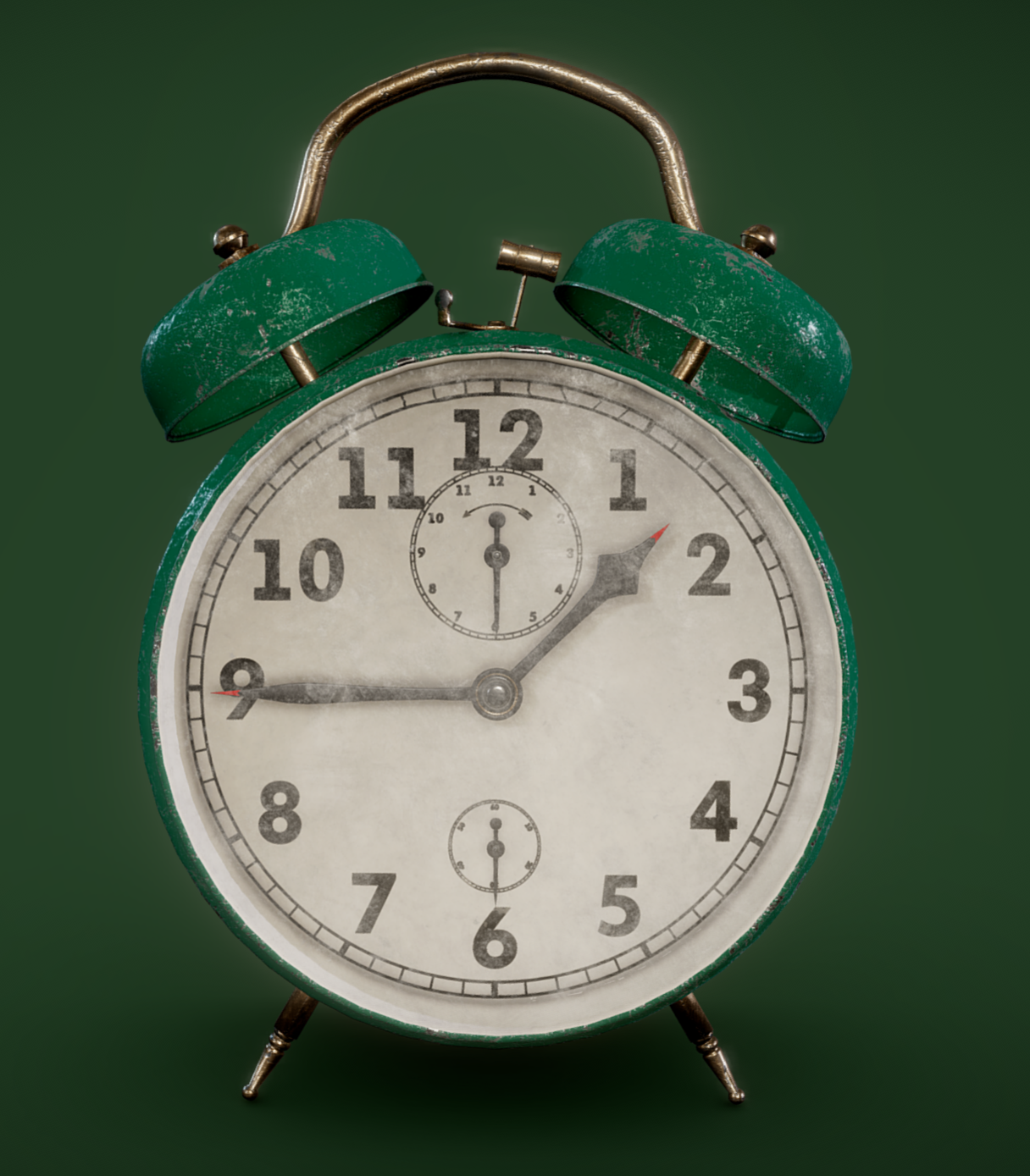}
        \end{tabular} &
        \textbf{Thinking}: The previous action adjusted the alarm hand instead of the main time hand. To set the correct time, I need to identify the appropriate knob. I will switch to the front view to check the hand positions before adjusting the other knob.

        \textbf{Action}: Switch\_view(params=[front]) \\
        
        \bottomrule
    \end{tabular}
    \caption{Typical examples of each error type on VGEBench. For actions that include coordinate parameters, we mark the corresponding locations with a red ``\textcolor{red}{$\times$}''. For \texttt{Switch\_view} actions, we display both the start view and the end view.}
    \label{tab:error_analysis}
\end{table*}

\begin{table*}[t]
\centering
\small
\begin{tabular}{lccccccccc}
\toprule
Model &
SR $\uparrow$ &
SSR $\uparrow$ &
SPL $\uparrow$ &
State-F1 $\uparrow$ &
EIR $\uparrow$ &
TIR $\uparrow$ &
VSPS $\downarrow$ &
GVPE $\downarrow$ &
EER $\uparrow$ \\
\midrule
Gemini-3-Flash &
56.38 & 64.32 & 39.23 & 78.51 &
64.79 & 42.77 & 1.18 & 1.94 & 63.11 \\
Human Baseline &
\textbf{74.50} & \textbf{81.43} & \textbf{54.86} & \textbf{85.65} &
\textbf{80.22} & \textbf{52.89} & \textbf{1.12} & \textbf{1.36} & \textbf{75.72} \\
\bottomrule
\end{tabular}
\caption{Complete human baseline results on 149 episodes. Human participants and Gemini-3-Flash are evaluated under the same interface and interaction budgets.}
\label{tab:human_full}
\end{table*}

\begin{table*}[t]
\centering
\small
\begin{tabular}{lrrr}
\toprule
Error Type & Gemini-3-Flash & Human Baseline & Difference (pp) \\
\midrule
Invalid Coord Error       & 35.5\% & 26.2\% & $-9.3$ \\
Component Selection Error & 30.6\% & 36.8\% & $+6.2$ \\
Params Error              & 10.8\% & 10.5\% & $-0.3$ \\
Action Error              &  4.9\% &  4.5\% & $-0.4$ \\
Invalid Waiting           &  1.5\% &  1.8\% & $+0.3$ \\
Switch View NN            & 12.1\% & 15.4\% & $+3.3$ \\
Switch View PN            &  4.6\% &  4.8\% & $+0.2$ \\
\bottomrule
\end{tabular}
\caption{Comparison of error distributions between Gemini-3-Flash and the human baseline. Percentages denote the proportion of each error type among all observed errors for the corresponding evaluator. Difference is computed as Human Baseline minus Gemini-3-Flash in percentage points.}
\label{tab:human_error}
\end{table*}

\section{Human Baseline Analysis}
\label{sec:human_baseline}

To contextualize current VLM performance, we evaluate a human baseline on 149 episodes (approximately 1\% of VGEBench) under the same interface and interaction budgets as Gemini-3-Flash.

Table~\ref{tab:human_full} reports the complete results across all evaluation metrics. Humans consistently outperform Gemini-3-Flash across the major task-performance, visual grounding, and exploration metrics. Human performance nevertheless remains below saturation, indicating that manual-free operation of unfamiliar devices remains non-trivial even for humans under the constrained interaction setting.

\paragraph{Error Distribution Comparison.}
We further compare the distributions of interaction errors between humans and Gemini-3-Flash using the same error taxonomy introduced in Appendix~\ref{sec:error_analysis}. As shown in Tab.~\ref{tab:human_error}, Humans and Gemini-3-Flash show similar overall distributions of Action Execution Errors and Visual Navigation Errors, with the main differences concentrated in Fine-grained Visual Grounding Errors: the Invalid Coord Error rate for humans is 26.2\%, which is 9.3 percentage points lower than that of Gemini-3-Flash, while the Component Selection Error rate is 36.8\%, which is 6.2 percentage points higher.

This pattern can be explained from the perspective of the information gained during interaction. Humans make fewer invalid coordinate errors, indicating that \textbf{humans can localize actual interactive regions more precisely, thereby triggering valid component-level feedback and completing a more informative and effective hypothesis-interaction-refinement loop under the same interaction budget}. In contrast, the model more frequently selects invalid coordinates, causing the feedback to remain at the lower-information “invalid coordinate” level and preventing it from fully entering subsequent functional inference and feedback-driven refinement. Therefore, the human baseline not only shows that the model continues to lag behind humans, but also provides supporting evidence for the importance of fine-grained visual grounding to exploration efficiency.

\section{Environment Mechanism Details}
\label{sec:appendix_environment}

\subsection{Environment Feedback Mechanism}
\label{sec:appendix_feedback}
The environment feedback is generated deterministically based on the matching result between the agent's action and the Specific State Machine (SSM). We categorize the feedback into three distinct levels to guide the agent's error recovery:
\begin{itemize}
    \item \textbf{Successful Transition:} If a transition occurs, the system returns the description of the transition and the new state (e.g., ``Power connected, the screen has turned on''). 
    \item \textbf{Operational Errors:} If the agent interacts with a valid component but uses an incorrect action or parameters (e.g., rotating a button that should be pushed), the system provides a hint: ``It seems that this component can be operated, but not in this way.''
    \item \textbf{Null Effect:} If the coordinates do not intersect with any interactive region, the system returns a ``null'' feedback.
\end{itemize}

\subsection{Interaction Budget Configuration}
\label{sec:appendix_budget}
To ensure efficient task execution, the episode terminates if all sub-tasks are completed or if any of the two budget constraints is violated:

\textbf{Global Interaction Budget ($B_{\text{global}}$):} To limit the absolute resource consumption, the total number of interaction steps across the entire episode must not exceed a static upper bound $B_{\text{global}}$. This bound is determined by the total complexity of the task, defined as
\begin{equation}
    B_{\text{global}} = \lambda \cdot \sum C_{\text{opt}},
\end{equation}
where $\sum C_{\text{opt}}$ represents the cumulative optimal cost across all sub-goals. Here, the optimal cost function is calculated as
\begin{equation}
    C_{\text{opt}} = d(S_{\text{start}}, S_{\text{end}}) + k_{\text{view}},
\end{equation}
composed of the shortest transition distance $d(\cdot)$ on the ground-truth state machine graph and the minimum number of view switches $k_{\text{view}}$ required along the path. $\lambda$ is a \textit{redundancy factor} defined to tolerate reasonable exploration, which is set to $5$ in our main experiments.

\textbf{Local Interaction Budget ($B_{\text{local}}$):} To prevent inefficient looping within specific sub-tasks, we maintain a dynamic interaction budget $B_{\text{local}}$ for the current execution phase. $B_{\text{local}}$ is initialized to a base value and decrements by 1 at each step. When the agent successfully completes a sub-goal and proceeds to the next target state $S_{\text{next}}$, $B_{\text{local}}$ is adaptively replenished based on the topological complexity of the upcoming task:
\begin{equation}
    B_{\text{local}} \leftarrow \max(B_{\text{local}}, \lambda \cdot C_{\text{opt}}).
\end{equation}

This mechanism ensures that the local budget is dynamically aligned with the difficulty of the current sub-task.

This dual-constraint mechanism enforces a rigorous standard for efficiency: the \textit{Global Constraint} ensures the agent manages resources over the long horizon, while the \textit{Dynamic Local Constraint} demands that the agent makes progress on specific sub-tasks without getting stuck in local optima. By grounding the budget in the ground-truth state machine ($C_{\text{opt}}$), the protocol ensures the allowance is always proportional to the actual task difficulty.



\subsection{View Mechanism}
\label{sec:appendix_view}

The views include six orthographic views (\texttt{front}, \texttt{back}, \texttt{left}, \texttt{right}, \texttt{top}, \texttt{bottom}), two axonometric views (\texttt{front\_top}, \texttt{back\_bottom}) and a special \texttt{``dashboard''} view. Because common rectangular buttons are prone to distortion in axonometric views, which makes bounding box annotation difficult, we restrict the agent’s interactions to the orthographic views. The axonometric views are only used to provide the agent with global visual information.

\begin{figure}[ht]
  \centering
  \includegraphics[width=\linewidth]{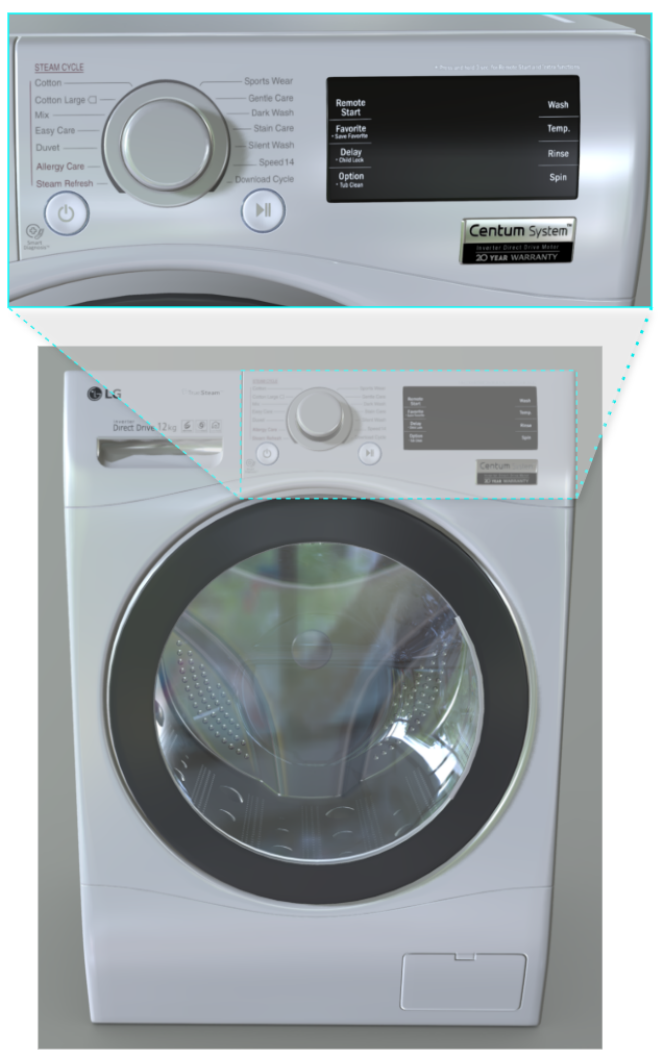}
  \caption{Illustration of the Dashboard View mechanism. For large devices where descriptive text or buttons are too small relative to the global view, this specialized view simulates the human perspective of approaching the object for closer inspection, enabling precise fine-grained interactions.}
  \label{fig:dashboard}
\end{figure}

\paragraph{Dashboard View Mechanism.}
For large household devices such as washing machines, the descriptive text or buttons are often too small relative to the device’s global view, making interactive components difficult to recognize in the standard view. To address this challenge, we introduce a specialized \textbf{Dashboard View} that simulates the human perspective of approaching an object for closer inspection, as illustrated in Fig.~\ref{fig:dashboard}. To prevent the agent from taking shortcuts, we do not provide the Dashboard View directly. Instead, we design two distinct unlocking mechanisms:

\textbf{Active Unlock:} Triggered when the agent explicitly executes a Switch\_view command with target coordinates falling within the dashboard region, resulting in an immediate transition to the Dashboard View.

\textbf{Passive Unlock:} Triggered implicitly when the agent attempts a spatial interaction (e.g., \texttt{Press}) within the locked region. In this case, the intended action is aborted and the view remains unchanged, but the dashboard state is updated to “unlocked” for subsequent operations.

Once unlocked, the dashboard functions as a standard view, enabling precise fine-grained interactions.

\subsection{Temporal and Chain Transitions}
\label{sec:appendix_transitions}
The simulator supports non-atomic state changes to model complex device behaviors:

\textbf{Immediate Transitions:} The state machine supports ``immediate'' nodes. If the system transitions to a state marked as immediate, it automatically evaluates the next transition without waiting for agent input. This allows for causal chains to be represented as a single step to the user.

\textbf{Timed Actions:} To support processes that require duration (e.g., a machine booting up), the system includes a \texttt{Timed\_wait} action. This action is only valid when the current state defines a temporal transition; otherwise, it returns feedback indicating that waiting had no effect.

\section{Additional Analysis}
\label{sec:additional-analysis}

In this section, we conduct analysis experiments on a randomly sampled 10\% subset of VGEBench.

\subsection{History Length Sensitivity}
\label{sec:history-sensitivity}

We vary the history length provided to the agent during multi-turn interaction, where ``No Limit'' denotes access to the complete interaction history, and smaller values indicate that only the most recent interaction turns are retained.

\begin{table}[h]
\centering
\small
\begin{tabular}{lcccc}
\toprule
\multirow{2}{*}{History Length} 
& \multicolumn{2}{c}{Gemini-3-Flash} 
& \multicolumn{2}{c}{Doubao-1.5} \\
\cmidrule(lr){2-3} \cmidrule(lr){4-5}
& SR & SSR & SR & SSR \\
\midrule
No Limit & 55.04 & 63.29 & 16.30 & 24.15 \\
10       & 51.10 & 59.88 & 18.19 & 25.99 \\
3        & 48.70 & 58.29 & 18.66 & 26.21 \\
1        & 42.47 & 51.69 & 17.19 & 24.58 \\
0        & 22.61 & 31.83 & 13.24 & 20.26 \\
\bottomrule
\end{tabular}
\caption{The result of history length sensitivity analysis. ``No Limit'' denotes that the complete interaction history is provided to the agent.}
\label{tab:history_length_sensitivity}
\end{table}

As shown in Table~\ref{tab:history_length_sensitivity}, Gemini-3-Flash consistently benefits from longer interaction history. Its SR and SSR increase from 22.61\% and 31.83\% with no history to 55.04\% and 63.29\% when the complete history is retained. This suggests that Gemini-3-Flash can effectively use long-horizon interaction traces to accumulate evidence and perform self-correction. In contrast, Doubao-1.5-Thinking-Vision-Pro achieves its best performance with a shorter history window of 3 turns, slightly outperforming the no-limit setting. This indicates that long histories may introduce distracting information for weaker models, especially when early failed attempts or irrelevant visual states remain in the context. Overall, the results show that interaction history is not merely a context-length variable, but also affects how different models balance useful feedback against historical noise.

\subsection{Visual Perturbation Robustness}
\label{sec:visual-perturbation}

Following ImageNet-C~\cite{hendrycks2019robustness} and Random Erasing~\cite{zhong2020random}, we apply three common visual perturbations to the input observations: image compression, Gaussian blur, and random occlusion. We designed the severity levels to ensure equal intervals and observable gradients:

\begin{itemize}
    \item \textbf{Compression (Image Quality):} We tested quality levels at \textbf{50} (mild artifacts) and \textbf{20} (severe blocking), simulating low-bandwidth transmission or low-quality sensors.
    \item \textbf{Blur (Gaussian):} We applied Gaussian blur with \textbf{$\sigma \in \{1.0, 3.0, 5.0\}$}, simulating motion blur or defocusing.
    \item \textbf{Occlusion:} We employed a ``Coarse Dropout'' strategy~\cite{zhong2020random} with 8 random black patches, covering total area ratios of \textbf{10\%}, \textbf{25\%}, and \textbf{40\%}, to simulate environmental clutter blocking the view.
\end{itemize}

Tables~\ref{tab:visual_perturbation_gemini} and~\ref{tab:visual_perturbation_doubao} show that both models are particularly sensitive to Gaussian blur. For Gemini-3-Flash, SR drops from 55.04\% to 38.53\% under strong blur with $\sigma=5.0$, and SSR decreases from 63.29\% to 48.12\%. Doubao-1.5-Thinking-Vision-Pro shows a similar trend, with SR decreasing from 16.30\% to 10.16\% under the same blur severity. In contrast, compression and occlusion lead to more moderate degradation. These results further support our main finding that fine-grained visual grounding is a central bottleneck in VGEBench: when high-frequency details such as small component boundaries, icons, and textual cues are degraded, models struggle to maintain reliable interaction performance.

\begin{table}[t]
\centering
\small
\begin{tabular}{llcc}
\toprule
Perturbation Type & Severity Level & SR & SSR \\
\midrule
Baseline                   & None          & 55.04 & 63.29 \\
\midrule
\multirow{2}{*}{Compression} & Quality = 50  & 53.29 & 62.83 \\
                           & Quality = 20  & 51.12 & 60.16 \\
\midrule
\multirow{3}{*}{Blur}        & $\sigma=1.0$  & 51.85 & 60.94 \\
                           & $\sigma=3.0$  & 40.92 & 51.18 \\
                           & $\sigma=5.0$  & 38.53 & 48.12 \\
\midrule
\multirow{3}{*}{Occlusion}   & Ratio = 0.10  & 50.14 & 59.44 \\
                           & Ratio = 0.25  & 50.73 & 58.69 \\
                           & Ratio = 0.40  & 47.05 & 56.72 \\
\bottomrule
\end{tabular}
\caption{The result of visual perturbation robustness analysis of Gemini-3-Flash.}
\label{tab:visual_perturbation_gemini}
\end{table}

\begin{table}[t]
\centering
\small
\begin{tabular}{llcc}
\toprule
Perturbation Type & Severity Level & SR & SSR \\
\midrule
Baseline                     & None          & 16.30 & 24.15 \\
\midrule
\multirow{2}{*}{Compression} & Quality = 50  & 16.88 & 25.22 \\
                             & Quality = 20  & 16.31 & 23.27 \\
\midrule
\multirow{3}{*}{Blur}        & $\sigma=1.0$  & 14.65 & 22.56 \\
                             & $\sigma=3.0$  & 10.28 & 17.40 \\
                             & $\sigma=5.0$  & 10.16 & 15.60 \\
\midrule
\multirow{3}{*}{Occlusion}   & Ratio = 0.10  & 15.94 & 22.97 \\
                             & Ratio = 0.25  & 17.23 & 25.83 \\
                             & Ratio = 0.40  & 14.57 & 22.57 \\
\bottomrule
\end{tabular}
\caption{The result of visual perturbation robustness of Doubao-1.5-Thinking-Vision-Pro.}
\label{tab:visual_perturbation_doubao}
\end{table}

\subsection{Cross-Model Instruction Robustness}
\label{sec:cross_model_instruction}

\begin{table}[t]
\centering
\small
\begin{tabular}{llcc}
\toprule
Generator & Evaluated Model & SR & SSR \\
\midrule
GPT-5-mini      & \multirow{3}{*}{Gemini-3-Flash} & 55.04 & 63.29 \\
DeepSeek-v4-pro &                                 & 55.52 & 63.38 \\
Qwen3.5-Plus    &                                 & 56.86 & 64.19 \\
\midrule
GPT-5-mini      & \multirow{3}{*}{GPT-5-mini} & 12.58 & 19.20 \\
DeepSeek-v4-pro &                             & 10.71 & 17.20 \\
Qwen3.5-Plus    &                             & 11.04 & 18.05 \\
\bottomrule
\end{tabular}
\caption{The result of cross-model instruction robustness analysis on a randomly sampled 10\% subset of VGEBench.}
\label{tab:cross_model_instruction}
\end{table}

To examine whether the evaluation is sensitive to the choice of instruction generator, we regenerate the task instructions using two additional LLMs, DeepSeek-v4-pro and Qwen3.5-Plus. We evaluate both Gemini-3-Flash and GPT-5-mini using instructions generated by GPT-5-mini, DeepSeek-v4-pro, and Qwen3.5-Plus.

As shown in Tab.~\ref{tab:cross_model_instruction}, changing the instruction generator results in only minor performance variations for both evaluated models. Gemini-3-Flash achieves SR between 55.04\% and 56.86\% and SSR between 63.29\% and 64.19\%, while GPT-5-mini achieves SR between 10.71\% and 12.58\% and SSR between 17.20\% and 19.20\%. In particular, using GPT-5-mini-generated instructions to evaluate GPT-5-mini does not produce a disproportionate performance increase compared with instructions generated by the other models. These results indicate that the evaluation is robust to the choice of instruction generator and does not exhibit a strong generator-specific performance advantage.

\subsection{Instruction Style Robustness}
\label{sec:instruction-style-robustness}

To address the concern that the current fixed language pipeline might introduce template and stylistic biases, we design an additional experiment comparing multiple instruction rewriting prototypes. The results are reported in Table~\ref{tab:instruction_style_robustness}.

The experimental results demonstrate that our framework exhibits strong robustness to rewriting styles. The standard deviations across both metrics are consistently \textbf{below 0.6\%}, and the narrow confidence intervals (CIs) further confirm the stability of the relative model ordering.

\begin{table*}[th]
\centering
\small
\begin{tabular}{llcccccc}
\toprule
Model & Metrics & Version 1 & Version 2 & Version 3 & Version 4 & Std Dev & 95\% CI \\
\midrule
\multirow{2}{*}{Gemini-3-Flash} 
& SR  & 51.64 & 51.51 & 52.71 & 52.44 & 0.59 & [49.89, 54.26] \\
& SSR & 60.09 & 59.84 & 60.95 & 59.99 & 0.50 & [58.25, 62.19] \\
\midrule
\multirow{2}{*}{Doubao-1.5-Thinking-Vision-Pro} 
& SR  & 16.66 & 17.26 & 17.39 & 18.06 & 0.58 & [15.64, 19.04] \\
& SSR & 23.96 & 24.50 & 24.50 & 25.09 & 0.46 & [22.78, 26.25] \\
\bottomrule
\end{tabular}
\caption{The result of instruction style robustness analysis. Performance is evaluated across four different instruction rewriting versions.}
\label{tab:instruction_style_robustness}
\end{table*}

\section{Detailed Evaluation Metrics}
\label{sec:metrics_details}

In this section, we provide the rigorous definitions and calculation details for the metrics used in our main evaluation.

\subsection{Task Performance} 
We primarily measure the \textbf{Success Rate (SR)}, defined as the percentage of test episodes where the model successfully achieves the goal state. For better evaluating step-wise correctness in multi-turn tasks, we additionally report the \textbf{Sub-task Success Rate (SSR)} which calculates the ratio of successfully achieved sub-goals to the total number of required sub-goals within an episode. 

\subsection{Efficiency}
Following~\citet{anderson2018evaluation}, we evaluate execution efficiency using two metrics:

\paragraph{Success Weighted by Path Length (SPL):} A metric that penalizes successful but inefficient executions. It is calculated as
\begin{equation}
    \text{SPL} = S \times \frac{L_{opt}}{\max(L_{act}, L_{opt})},
\end{equation}
where $S$ indicates success (1 or 0), $L_{opt}$ is the length of the theoretical shortest path derived from the state machine, and $L_{act}$ is the model's actual path length.

\paragraph{State-F1 Score:} To measure the alignment between the model's trajectory and the optimal solution, we treat the sequence of visited states as a set and compute the F1 score between the model's state trajectory and the ground-truth optimal path.

\subsection{Visual Grounding and Perception} 
Inspired by RefCOCO~\cite{yu2016modeling}, we evaluate the model's visual interaction capabilities using four metrics:

\paragraph{Effective Interaction Rate (EIR):} Measures the proportion of actions where the output coordinates $(x, y)$ fall within the bounding box of \textit{any} interactive component. This indicates the model's fundamental ability to identify and interact with valid UI elements.

\paragraph{Target Interaction Rate (TIR):} Measures the precision of goal-oriented grounding. An interaction is deemed accurate only if the coordinates fall within the bounding box of a component that triggers a transition to a state with a strictly shorter shortest-path distance to the current goal.

\paragraph{View Switches Per Success (VSPS)} is an episode-level metric that calculates the number of view switches executed prior to a successful action, normalized by the number of goals achieved. VSPS isolates \textit{local navigation efficiency}, measuring how directly the model navigates to the target view when it correctly understands the task.

\paragraph{Global View Switches Per Episode (GVPE)} is a dataset-level metric computed as the total number of view switches across all test episodes divided by the total number of achieved goals. Unlike VSPS, GVPE incorporates the cost of invalid exploration in failed episodes. It serves as a \textit{macro-level indicator} of the exploration-to-success ratio.

\subsection{Exploration} 
Inspired by Webshop~\cite{yao2022webshop}, we assess the model's ability to explore devices' functionality effectively by defining the \textbf{Effective Exploration Rate (EER)}. We classify operations into \textit{valid} and \textit{invalid} categories. Invalid operations include repetitive actions, dangerous actions, and null-transition actions that trigger no state change. The EER is calculated as the ratio of valid operations to the total number of operation steps, reflecting the model's reasoning efficiency and safety awareness.

\section{Data Validation Details}
\label{sec:validation_criteria}

\subsection{Annotation Quality Control}

Our annotation process consists of two parts. First, annotators labeled the bounding boxes of all visible interactive components across 7,888 rendered views. Second, annotators annotated the trigger conditions for 15,861 state transition edges in the instantiated SSMs. Each transition trigger specifies the reactive coordinate region, the corresponding atomic action type, and the required action parameters. We use transition edges rather than components as the annotation unit because the same physical component may correspond to different state-dependent transitions or require different actions and parameters under different device states (e.g., \textit{pull} to open vs. \textit{push} to close).

We adopted a batch-level quality-control protocol for both annotation stages. For bounding-box annotation, annotators first completed an initial familiarization batch of approximately 100 views, all of which were manually reviewed. In regular annotation, each batch contained no fewer than 1,000 views; with a 10\% random sampling rate, at least 100 views were reviewed per batch. For transition-trigger annotation, each batch contained no fewer than 2,000 transition edges; therefore, at least 200 transition edges were reviewed per batch.

A batch was accepted only when its error rate was below 5\%. Otherwise, the batch was rejected and reassigned for correction. This protocol provides reliable quality control for both visual component annotation and state-machine trigger annotation.

\subsection{Dataset Quality Validation}

\begin{table}[ht]
\centering
\small
\setlength{\tabcolsep}{8pt}
\renewcommand{\arraystretch}{1.2}
\begin{tabular}{l ccc c}
\toprule
\textbf{Evaluator} & \textbf{VGA} & \textbf{LC} & \textbf{IA} & \textbf{Avg.} \\
\midrule
Gemini-3-Flash & 1.944 & 1.997 & 1.991 & 1.978 \\
GPT5-mini    & 1.906 & 1.964 & 1.970 & 1.947 \\
\midrule
Human  & 1.910 & 1.954 & 1.965 & 1.943 \\
\bottomrule
\end{tabular}
\caption{
Data validation results on the validation subset. 
Scores are evaluated on a 3-level scale (0/1/2) and reported as the mean value across all samples.
}
\label{tab:qa_results}
\end{table}

In addition to batch-level annotation quality control, we established a rigorous quality validation scoring criteria to ensure dataset reliability. We defined three specific metrics to assess the quality of the visual grounding, logical feasibility, and instruction alignment. The detailed scoring criteria are described as follows:

\textbf{Visual Grounding Accuracy (VGA)}: Measures whether the annotated bounding box (visual context) accurately identifies the specific physical UI element corresponding to the execution action.
\begin{itemize}[itemsep=1pt, topsep=2pt, parsep=0pt]
    \item \textbf{bad(0)}: The highlighted region is clearly incorrect, unrelated to the target element, or physically implausible.
    \item \textbf{fair(1)}: The highlighted region partially matches the target or presents slight ambiguity regarding the specific element.
    \item \textbf{good(2)}: The highlighted region clearly and correctly corresponds to the required trigger for the transition.
\end{itemize}

\textbf{Logical Consistency (LC)}: Evaluates whether the state transition and action sequence are physically feasible and logically sound for the specific device configuration.
\begin{itemize}[itemsep=1pt, topsep=2pt, parsep=0pt]
    \item \textbf{bad(0)}: The transition from the start state to the end state is impossible or clearly contradicts physical constraints.
    \item \textbf{fair(1)}: The transition is mostly logical but may involve minor practical issues or unlikely aspects.
    \item \textbf{good(2)}: The transition is fully logical, feasible, and consistent with real-world physical behavior.
\end{itemize}

\textbf{Instruction Alignment (IA)}: Assesses whether the symbolic state transition represents a valid subtask that effectively contributes to the user's overall high-level intent.
\begin{itemize}[itemsep=1pt, topsep=2pt, parsep=0pt]
    \item \textbf{bad(0)}: The transition is clearly unrelated to any part of the user's overall task.
    \item \textbf{fair(1)}: The transition is partially related or exhibits minor mismatches with the user's specific intent.
    \item \textbf{good(2)}: The transition is fully contained within the user's intended task and logically aligns with the execution plan.
\end{itemize}

\section{Details of Exploration framework}
\label{sec:method_prompt}

Our exploration framework is based on a basic \textbf{ReAct} paradigm~\cite{yao2023react}, which employs an iterative reasoning–action loop to guide interaction with the environment. Given visual observations, the agent first analyzes the scene to identify potential interaction targets and infer device functionality. It then executes a single atomic action grounded on this reasoning, receives environment feedback, and continues the exploration process until the task objective is achieved.

\begin{table*}[h!]
    \centering
    \begin{promptbox}
    You are an intelligent robotic agent capable of manipulating a wide variety of household devices based on visual inputs.

    \medskip
    \textbf{1. Output Format}\\
    You must structure your response strictly in the following two parts:

    \textbf{Thinking:}\\
    \hllogic{Analyze the visual information to identify the target interactable area}. Reason about the device functionality and the physics of the interaction (e.g., whether it is a button, a knob, a door, or a purely temporal process). If the required interaction target is not visible in the current view, or the task cannot yet be executed, reason about the appropriate non-physical action.

    \textbf{Action:}\\
    Execute the operation using a precise function call format.\\
    \hlfor{Format: \texttt{ActionName(coord=[x, y], params=[...])}}\\
    \hlcons{For actions that do not require a spatial target, omit \texttt{coord}}.

    \medskip
    \textbf{2. Coordinate System}
    \begin{itemize}
        \item \textbf{Origin (0,0):} Top-Left corner of the image.
        \item \textbf{X-axis:} Extends horizontally to the Right.
        \item \textbf{Y-axis:} Extends vertically Downwards.
        \item \textbf{Scale:} \hlcons{Normalized integer coordinates ranging from \textbf{0 to 1000}} (Top-Left: \texttt{[0, 0]}, Bottom-Right: \texttt{[1000, 1000]}).
    \end{itemize}

    \textbf{3. Atomic Action Space}\\
    Select the most appropriate action from the atomic operations below.

    \textbf{Press(coord, params)}
    \begin{itemize}
        \item \textbf{Description:} Interacting with buttons, touchscreens, switches, or triggers.
        \item \textbf{Params:} \texttt{[duration\_type]}
        \begin{itemize}
            \item \texttt{0}: Click (Standard press)
            \item \texttt{1}: Long Press (Hold for $>$1s)
            \item \texttt{2}: Double Click (Rapid toggle)
        \end{itemize}
    \end{itemize}

    \textbf{Rotate(coord, params)}
    \begin{itemize}
        \item \textbf{Description:} Turning knobs, dials, keys, or mechanical winders.
        \item \textbf{Params:} \texttt{[direction, extent]}
        \begin{itemize}
            \item \texttt{direction}: \texttt{0} (Clockwise), \texttt{1} (Counter-clockwise), \texttt{*} (Unconstrained or irrelevant)
            \item \texttt{extent}: Integer representing the number of discrete rotational steps (e.g., notches or detents). \texttt{*} (Rotation magnitude is unspecified).
        \end{itemize}
    \end{itemize}

    \textbf{Push(coord, params)}
    \begin{itemize}
        \item \textbf{Description:} Applying force to slide objects, close doors, or insert plugs.
        \item \textbf{Params:} \texttt{[direction]}
        \begin{itemize}
            \item \texttt{0}: Up, \texttt{1}: Down, \texttt{2}: Left, \texttt{3}: Right
            \item \texttt{4}: In / Forward (e.g., closing a door, pushing a drawer in, inserting a cable)
        \end{itemize}
    \end{itemize}

    \textbf{Pull(coord, params)}
    \begin{itemize}
        \item \textbf{Description:} Applying force to open doors, remove lids, or unplug cables.
        \item \textbf{Params:} \texttt{[direction]}
        \begin{itemize}
            \item \texttt{0}: Up, \texttt{1}: Down, \texttt{2}: Left, \texttt{3}: Right
            \item \texttt{4}: Out / Backward (e.g., opening a fridge, pulling a drawer out, unplugging)
        \end{itemize}
    \end{itemize}
    \end{promptbox}
    \caption{System Prompt Part 1: Output Format, Coordinate System and Action Space. \hlfor{Format constraint}, \hlcons{constraints \& boundaries}, \hllogic{logical mechanics} are highlighted.}
\end{table*}

\begin{table*}[h!]
    \centering
    \begin{promptbox}

    \textbf{Grasp(coord, params)}
    \begin{itemize}
        \item \textbf{Description:} Holding a handle or gripping a handheld device.
        \item \textbf{Params:} \texttt{[state]} (\texttt{0}: Hold / Pick up, \texttt{1}: Release / Put down)
    \end{itemize}

    \textbf{Switch\_view(params)}
    \begin{itemize}
        \item \textbf{Description:} Switching to a different camera angle or viewpoint when the required interaction target is not visible in the current view.
        \item \textbf{Params:} \texttt{[view]} (Symbolic view identifier selected from optional view list).
        \item Common views include six standard views (\texttt{front}, \texttt{back}, \texttt{left}, \texttt{right}, \texttt{top}, \texttt{bottom}), two axonometric views (\texttt{front\_top}, \texttt{back\_bottom}), and a special \texttt{dashboard} view. \hlcons{You cannot perform any actions} \hlcons{except \texttt{Switch\_view} when the current view is an axonometric view}.
        \item \textbf{Dashboard View:} Some devices provide a \texttt{dashboard} view for interfaces where text or buttons are dense relative to the overall device size.
        If listed as \texttt{dashboard(locked)} in optional view list, it must be unlocked before use.

        Unlocking methods:
        \begin{enumerate}
            \item \textbf{Active unlock:}
            \hlfor{\texttt{Switch\_view(coord=[x, y], params=[dashboard])}}
            If \texttt{coord} falls within the dashboard region, the dashboard is unlocked and the view switches immediately.
            \item \textbf{Passive unlock:}
            Using another action (e.g., \texttt{Press}) with \texttt{coord} inside the dashboard region also unlocks the dashboard.
            \hlcons{In this case, the current action does not execute and the} \hlcons{view does not switch}.
        \end{enumerate}

        Once unlocked, \texttt{dashboard} behaves like a standard view. You are encouraged to perform interactions directly within the dashboard view when available.
    \end{itemize}

    \textbf{Timed\_wait (params)}
    \begin{itemize}
        \item \textbf{Description:} Waiting for a time-dependent process to complete before further interaction.
        \item \textbf{Params:} \texttt{[]} (\hlfor{No parameters are required}).
    \end{itemize}

    \textbf{4. Constraint:}
    \hlcons{The agent must output exactly \textbf{one atomic action per response}}. Each response may contain only one ``Action: '' entry, corresponding to a single atomic operation from the defined action space.

    \textbf{Example Output}

    \textbf{Example 1}\\
    Thinking: The user wants to change the mode to ``Quick''. I see a rotary dial labeled ``Mode'' at the bottom right. The current mode is ``Medium'', and switching to ``Quick'' requires rotating the dial two discrete levels clockwise. I will rotate it accordingly to fulfill the user's request.\\
    Action: \texttt{Rotate(coord=[850, 720], params=[0, 2])}

    \textbf{Example 2}\\
    Thinking: The user wants to charge the phone. I cannot find a charging port in the current view, which suggests it may be located on another side of the device. Since charging ports are typically located at the bottom of smartphones, I should switch to the bottom view to continue the task.\\
    Action: \texttt{Switch\_view(params=[bottom])}

    \textbf{Example 3}\\
    Thinking: The user wants to take out the toasted bread. The machine is still operating, and the toasting process has not finished yet. I need to wait until the operation completes before performing any further interaction.\\
    Action: \texttt{Timed\_wait()}

    \textbf{Example 4}\\
    Thinking: The user wants to change the mode to ``Quick''. I see a rotary dial, but I cannot clearly recognize the labels next to it. I should switch to the dashboard view to inspect the controls more closely.\\
    Action: \texttt{Switch\_view(coord=[850, 720], params=[dashboard])}

    \end{promptbox}
    \caption{System Prompt Part 2: Action Space, Constraint and Examples.  \hlfor{Format constraint}, \hlcons{constraints \& boundaries} are highlighted.}
    \label{tab:react_system_prompt}
\end{table*}

\begin{table*}[h]
    \centering
    \begin{promptbox}
    You are a helpful assistant generating user commands for interacting with a household device.

    \medskip
    \textbf{Input Data}
    \begin{itemize}
        \item \textbf{Device Type:} \{device\_type\}
        \item \textbf{State Transitions Sequence:} \{transition\_display\}
        \item \textbf{Context Information:} \{context\_info\} (Detailed descriptions of states and actions involved)
    \end{itemize}

    \textbf{Task}\\
    Convert this sequence of state transitions into a list of natural language user commands.
    \hlcons{There should be exactly one command for each transition step}.

    \medskip
    \textbf{Requirements}
    \begin{enumerate}
        \item \hlfor{Output a JSON list of strings}, e.g., \texttt{["Command for step 1", "Command for step 2"]}.
        \item The commands should be imperative and natural.
        \item \hllogic{If it makes sense, you can add natural flow or connection between steps} (e.g., ``The wind is not strong enough, switch to high mode''), but ONLY if the transition logic supports it. Otherwise, faithfully describe the transition.
        \item \hlcons{Avoid explicitly mentioning button names (like ``power button''), focus on function/intent}.
        \item \hlcons{Do not output anything other than the JSON list}.
    \end{enumerate}

    \medskip

    \textbf{Examples}

    \textbf{Example 1}
    \begin{itemize}
        \item \textbf{Transitions:} \texttt{["Off->Active", "Active->High"]}
        \item \textbf{Output:} \texttt{["Turn on the flashlight", "The light is too dim, switch to high mode"]}
    \end{itemize}

    \textbf{Example 2}
    \begin{itemize}
        \item \textbf{Transitions:} \texttt{["Standby->Running"]}
        \item \textbf{Output:} \texttt{["Start the washing cycle"]}
    \end{itemize}

    \end{promptbox}
    \caption{Prompt for Generating User Commands from State Transitions. \hlfor{Format constraint}, \hlcons{constraints \& boundaries}, \hllogic{logical mechanics} are highlighted.}
    \label{tab:command_generation_prompt}
\end{table*}

\clearpage
\onecolumn
\twocolumn

\section{Data Sample}
\label{sec:data_sample}

In this section, we present a detailed example of the state machine structure used in VGEBench, taking the \textit{Electric Kettle} as a case study. We illustrate both the Universal Category State Machine (UCSM) and the derived Specific State Machine (SSM).

\subsection{Data Structure}
Each device entry in our dataset consists of three primary fields: \texttt{name}, \texttt{initial\_state}, and \texttt{states}. The \texttt{states} field is a dictionary that defines the topology of the logic graph. Each state object includes:
\begin{itemize}
    \item \texttt{description}: A natural language description of the current device status.
    \item \texttt{on}: A dictionary defining valid transitions. The keys correspond to \textbf{interactive components} (e.g., \textit{power\_button}), and the values are lists of transition objects. A list is used because interacting with a single component can lead to different states depending on the action parameters (e.g., rotating a knob to position 1 vs. position 2).
\end{itemize}

\paragraph{Universal Category State Machine (UCSM).} 
In the UCSM, each transition object primarily defines the \texttt{next\_state} and a \texttt{transition\_description}. 
Notably, some UCSMs include a special \texttt{Error} state. This state captures dangerous or invalid operations (e.g., opening the lid while the device is running). If an agent triggers such a transition, the device enters the \texttt{Error} state and is subsequently reset to the \texttt{initial\_state}, simulating a safety mechanism.

\paragraph{Specific State Machine (SSM).} 
The SSM is derived from the UCSM by pruning transitions associated with components that are not visually present on the specific 3D model instance. Unreachable states and transitions are subsequently removed to ensure graph connectivity.
Crucially, the SSM enriches each transition with ground-truth execution details:
\begin{itemize}
    \item \texttt{bbox}: A list of bounding boxes for the interactive component. This is a list because a single component may be visible and actionable from multiple viewpoints (e.g., \textit{front}, \textit{top}, \textit{right}).
    \item \texttt{atomic\_action}: The specific action type required (e.g., \texttt{push}, \texttt{rotate}).
    \item \texttt{parameter}: The precise parameters for the action (e.g., \texttt{["1"]}).
\end{itemize}

\subsection{JSON Examples}

Listing~\ref{lst:ucsm_example} shows the structure of the UCSM for the Electric Kettle category, and Listing~\ref{lst:ssm_example} displays the instantiated SSM for a specific kettle instance, featuring detailed visual grounding annotations. For data privacy reasons, the ``path'' field has been anonymized.

\onecolumn

\begin{lstlisting}[
    language=json,
    caption={Example of the Universal Category State Machine (UCSM) for Electric Kettle.},
    label={lst:ucsm_example}
]
{
  "name": "ElectricKettle",
  "initial_state": "Standby",
  "states": {
    "Off": {
      "description": "Disconnect the power cord; the electric kettle is powered off",
      "on": {
        "power_input_port": [
          {
            "next_state": "Standby",
            "transition_description": "Power connected, entering standby mode"
          }
        ]
      }
    },
    "Standby": {
      "description": "Standby state, power connected but not started",
      "on": {
        "power_button": [
          {
            "next_state": "Boiling",
            "transition_description": "Start the electric kettle; begin heating"
          }
        ],
        "lid_release_button": [
          {
            "next_state": "LidOpen",
            "transition_description": "Open the kettle lid and prepare to add water"
          }
        ],
        "temperature_control": [
          {
            "next_state": "StandbyTemperatureControl",
            "transition_description": "Adjust the keep-warm/target temperature setting"
          }
        ],
        "power_input_port": [
          {
            "next_state": "Off",
            "transition_description": "Unplug the power cord; enter a power-off state"
          }
        ]
      }
    },
    "StandbyTemperatureControl": {
      "description": "Adjusting keep-warm/target temperature settings",
      "on": {
        "immediate": [
          {
            "next_state": "Standby",
            "transition_description": "Temperature setting completed; return to standby state"
          }
        ]
      }
    },
    "Boiling": {
      "description": "Heating or boiling water",
      "on": {
        "timed_wait": [
          {
            "next_state": "BoilingComplete",
            "transition_description": "Water has boiled"
          }
        ],
        "power_button": [
          {
            "next_state": "Standby",
            "transition_description": "Press the power button to stop heating"
          }
        ],
        "temperature_control": [
          {
            "next_state": "BoilingTemperatureControl",
            "transition_description": "Adjust the keep-warm/target temperature during heating"
          }
        ],
        "power_input_port": [
          {
            "next_state": "Off",
            "transition_description": "Forced power-off caused heating interruption"
          }
        ],
        "lid_release_button": [
          {
            "next_state": "Error",
            "transition_description": "Forcibly opening the top cover during heating, dangerous operation"
          }
        ]
      }
    },
    "BoilingTemperatureControl": {
      "description": "Adjust the keep-warm/target temperature during heating",
      "on": {
        "immediate": [
          {
            "next_state": "Boiling",
            "transition_description": "Finished adjusting temperature; return to heating state"
          }
        ]
      }
    },
    "LidOpen": {
      "description": "Open the kettle lid and add water",
      "on": {
        "lid_release_button": [
          {
            "next_state": "Standby",
            "transition_description": "After adding water, close the kettle lid and return to standby"
          }
        ],
        "power_button": [
          {
            "next_state": "Error",
            "transition_description": "Attempting to heat with the lid open, erroneous operation"
          }
        ]
      }
    },
    "BoilingComplete": {
      "description": "Water has boiled, boiling process completed",
      "on": {
        "immediate": [
          {
            "next_state": "Standby",
            "transition_description": "Pour out the brewed water and return to standby"
          }
        ]
      }
    },
    "Error": {
      "description": "Error operation state",
      "on": {
        "immediate": [
          {
            "next_state": "Standby",
            "transition_description": "An incorrect operation occurred; automatically reset to the initial state"
          }
        ]
      }
    }
  }
}
\end{lstlisting}

\begin{lstlisting}[
    language=json,
    caption={Example of the Specific State Machine (SSM) for a specific kettle instance. Transitions now include fine-grained bounding boxes across multiple views, atomic actions and corresponding parameters.},
    label={lst:ssm_example}
]
{
  "name": "ElectricKettle",
  "path": "image_view\\ElectricKettle\\device number",
  "initial_state": "Standby",
  "states": {
    "Standby": {
      "description": "Standby state, power connected but not started",
      "on": {
        "power_button": [
          {
            "next_state": "Boiling",
            "transition_description": "Start the electric kettle; begin heating",
            "bbox": [
              {
                "coordinate": [199, 815, 143, 57, 1138, 1122],
                "view": "back"
              },
              {
                "coordinate": [915, 373, 160, 135, 1195, 908],
                "view": "bottom"
              },
              {
                "coordinate": [852, 955, 184, 76, 1209, 1369],
                "view": "front"
              },
              {
                "coordinate": [324, 978, 135, 66, 786, 1191],
                "view": "right"
              }
            ],
            "atomic_action": "push",
            "parameter": ["1"]
          }
        ],
        "lid_release_button": [
          {
            "next_state": "LidOpen",
            "transition_description": "Open the kettle lid and prepare to add water",
            "bbox": [
              {
                "coordinate": [216, 61, 98, 27, 1138, 1122],
                "view": "back"
              },
              {
                "coordinate": [881, 81, 117, 34, 1209, 1369],
                "view": "front"
              },
              {
                "coordinate": [336, 87, 84, 44, 786, 1191],
                "view": "right"
              },
              {
                "coordinate": [1005, 377, 126, 96, 1292, 760],
                "view": "top"
              }
            ],
            "atomic_action": "push",
            "parameter": ["1"]
          }
        ]
      }
    },
    "Boiling": {
      "description": "Heating or boiling water",
      "on": {
        "timed_wait": [
          {
            "next_state": "BoilingComplete",
            "transition_description": "Water has boiled"
          }
        ],
        "power_button": [
          {
            "next_state": "Standby",
            "transition_description": "Press the power button to stop heating",
            "bbox": [
              {
                "coordinate": [199, 815, 143, 57, 1138, 1122],
                "view": "back"
              },
              {
                "coordinate": [915, 373, 160, 135, 1195, 908],
                "view": "bottom"
              },
              {
                "coordinate": [852, 955, 184, 76, 1209, 1369],
                "view": "front"
              },
              {
                "coordinate": [324, 978, 135, 66, 786, 1191],
                "view": "right"
              }
            ],
            "atomic_action": "push",
            "parameter": ["0"]
          }
        ],
        "lid_release_button": [
          {
            "next_state": "Error",
            "transition_description": "Forcibly opening the top cover during heating, dangerous operation",
            "bbox": [
              {
                "coordinate": [216, 61, 98, 27, 1138, 1122],
                "view": "back"
              },
              {
                "coordinate": [881, 81, 117, 34, 1209, 1369],
                "view": "front"
              },
              {
                "coordinate": [336, 87, 84, 44, 786, 1191],
                "view": "right"
              },
              {
                "coordinate": [1005, 377, 126, 96, 1292, 760],
                "view": "top"
              }
            ],
            "atomic_action": "push",
            "parameter": ["1"]
          }
        ]
      }
    },
    "LidOpen": {
      "description": "Open the kettle lid and add water",
      "on": {
        "lid_release_button": [
          {
            "next_state": "Standby",
            "transition_description": "After adding water, close the kettle lid and return to standby",
            "bbox": [
              {
                "coordinate": [216, 61, 98, 27, 1138, 1122],
                "view": "back"
              },
              {
                "coordinate": [881, 81, 117, 34, 1209, 1369],
                "view": "front"
              },
              {
                "coordinate": [336, 87, 84, 44, 786, 1191],
                "view": "right"
              },
              {
                "coordinate": [1005, 377, 126, 96, 1292, 760],
                "view": "top"
              }
            ],
            "atomic_action": "push",
            "parameter": ["0"]
          }
        ],
        "power_button": [
          {
            "next_state": "Error",
            "transition_description": "Attempting to heat with the lid open, erroneous operation",
            "bbox": [
              {
                "coordinate": [199, 815, 143, 57, 1138, 1122],
                "view": "back"
              },
              {
                "coordinate": [915, 373, 160, 135, 1195, 908],
                "view": "bottom"
              },
              {
                "coordinate": [852, 955, 184, 76, 1209, 1369],
                "view": "front"
              },
              {
                "coordinate": [324, 978, 135, 66, 786, 1191],
                "view": "right"
              }
            ],
            "atomic_action": "push",
            "parameter": ["1"]
          }
        ]
      }
    },
    "BoilingComplete": {
      "description": "Water has boiled, boiling process completed",
      "on": {
        "immediate": [
          {
            "next_state": "Standby",
            "transition_description": "Pour out the brewed water and return to standby"
          }
        ]
      }
    },
    "Error": {
      "description": "Error operation state",
      "on": {
        "immediate": [
          {
            "next_state": "Standby",
            "transition_description": "An incorrect operation occurred; automatically reset to the initial state"
          }
        ]
      }
    }
  }
}
\end{lstlisting}

\end{document}